\documentclass[letterpaper, 10 pt, conference]{ieeeconf}

\IEEEoverridecommandlockouts

\usepackage{cite}
\usepackage{amsmath,amssymb,amsfonts}
\usepackage{algorithmic}
\usepackage{graphicx}
\usepackage{textcomp}
\usepackage{xcolor}
\def\BibTeX{{\rm B\kern-.05em{\sc i\kern-.025em b}\kern-.08em
    T\kern-.1667em\lower.7ex\hbox{E}\kern-.125emX}}
    
\usepackage[ruled, vlined, linesnumbered]{algorithm2e}

\usepackage[labelformat=simple]{subcaption}
\DeclareCaptionLabelSeparator{periodspace}{.\quad}
\usepackage{tikz}
\usetikzlibrary{arrows,matrix,positioning,patterns}
\usetikzlibrary{calc}
\usepackage{pgfplots} 
\usetikzlibrary{plotmarks}
\usetikzlibrary{arrows.meta,calc}

\renewcommand{\baselinestretch}{0.905}
\usepackage{colortbl}

\begin{document}

\title{\LARGE \bf High-level spatial Dubins airplane-based reference smoothing with low-level geometric tracking for quadrotor control}


\author{Mogens Plessen*
\thanks{*{\tt\small pmogens@proton.me}, Findklein GmbH, Switzerland}
}

\maketitle

\begin{abstract}
A method for the control of quadrotors is presented. It is composed of a high-level reference smoothing step and a low-level smoothed reference tracking step. The high-level step leverages the Dubins airplane model for dimensionality reduction and reduced computational complexity, and exploits its structure for decoupling, spatial modeling and the formulation of a small linear program. The low-level step leverages a geometric tracking controller, which is based on the full quadrotor model. The method is designed for the tracking of references subject to lateral constraints along the path. An example is the tracking of references along obstacle contours. It is differentiated between two different setups. Either the high-level planning step is conducted once and offline, or, alternatively, the high-level planning step is conducted recedingly online in closed-loop over a limited spatial prediction horizon.
\end{abstract}


\section{Introduction}

\begin{figure}[htbp]
\centering
%
\begin{subfigure}[b]{0.245\textwidth}
\centering
\includegraphics[width=1.0\textwidth]{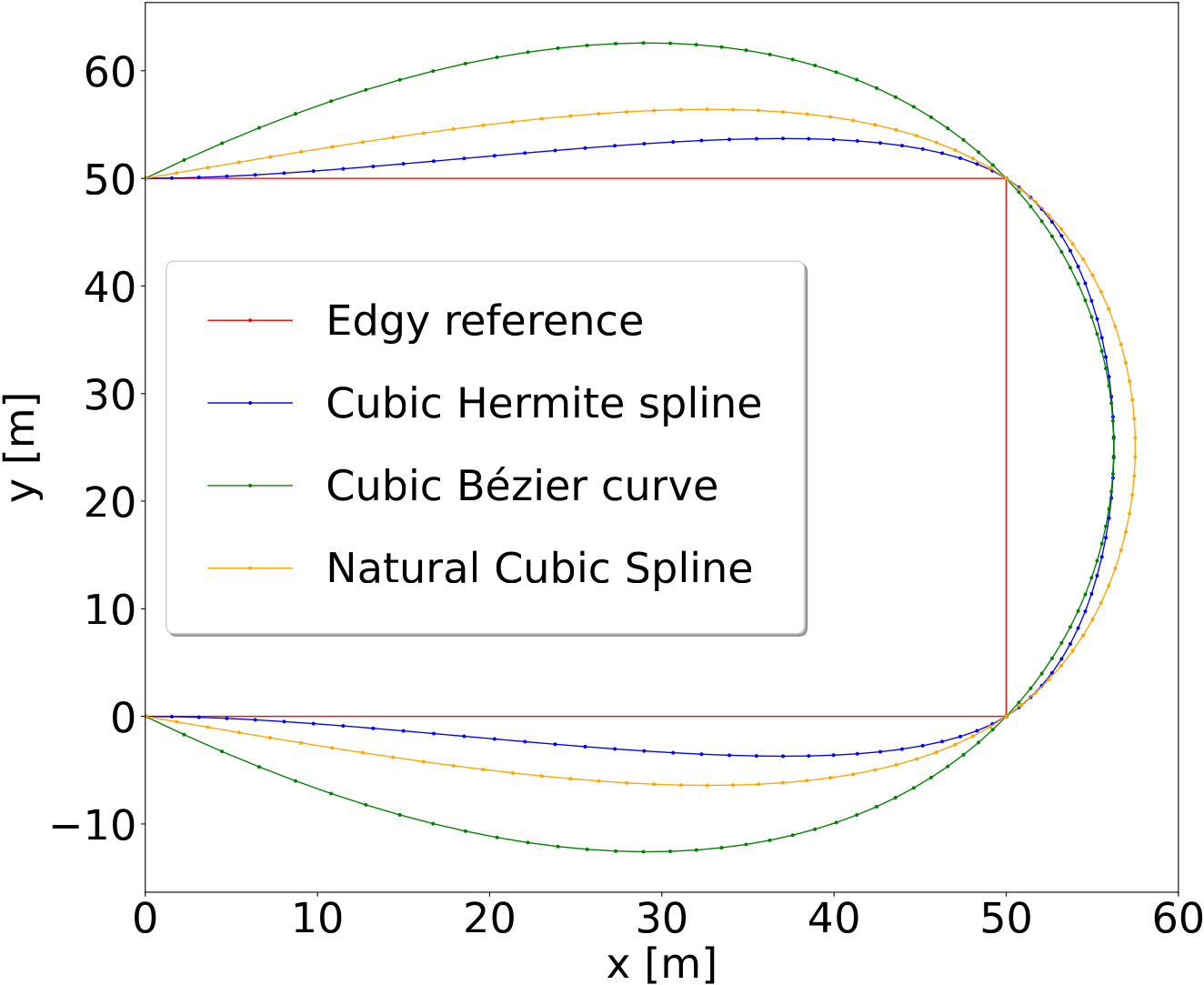}
\caption{Counterexample 1: different spline methods \cite{deboor2001splines} are used for smoothing of an edgy 4-waypoints reference. \vspace*{0.1cm}}
\label{fig_hermite}
\end{subfigure}
\begin{subfigure}[b]{0.23\textwidth}
\centering
\includegraphics[width=1.0\textwidth]{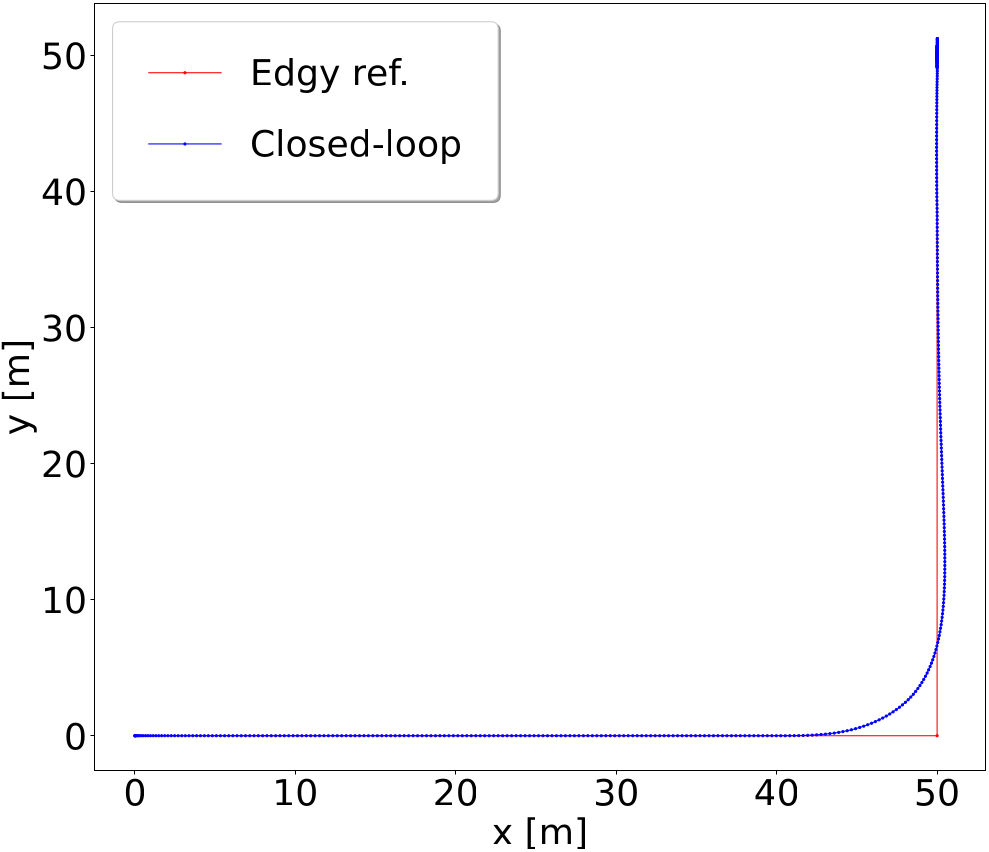}
\caption{Counterexample 2: a low-level controller is used for closed-loop tracking of a 3-waypoints reference.\vspace*{0.1cm}}
\label{fig_cornercut}
\end{subfigure}
%
\caption{Problem visualization by counterexamples: it is wished to (i) closely track an edgy sparse waypoints reference, (ii) while laterally staying on a specific side of the reference (e.g., avoiding the obstacle area left of the reference). Three spline-based methods fail condition (i) by staying far from the reference, while the method of directly tracking the edgy reference with a low-level controller fails condition (ii) by corner-cutting.}
\label{fig_problvisualization}
\end{figure}

The literature for quadrotor or in general multirotor control is very rich \cite{kim2020comprehensive,lopez2023pid}. This underlines both relevance as well as complexity. Examples are given for illustration. First, in \cite{kamel2015fast} a fast nonlinear model predictive control method is proposed for reference tracking. However, to reduce  computational complexity on-board their prediction horizon is only 50ms. Furthermore, their objective function requires the selection of two positive semi-definite matrices, a non-negative scalar and one positive definite matrix as hyperparameters, all in different units. Numerical values are not stated to evaluate required tuning effort. Similarly, in \cite{krinner2024mpcc++} the objective function of their optimal control problem,  which is solved off-board, consists of 6 different weighted terms, the prediction span is 0.8s, and a hermitian spline fit is used as centerline for tracking. Second, in the smoothing step of \cite{zhou2019robust}, B-spline optimization is used, requiring the selection of control points, the degree of the spline, an objective function with 3 hyperparameters weighting smoothness, collision cost, soft limits on velocity and acceleration, multiple threshold parameters, and a nonlinear optimization solver. Third, in \cite{kondo2026sando} a spatiotemporal safe flight corridor generation method is proposed based on cubic Bézier curves. However, Bézier curves, and similarly other spline-based methods, are not suitable for every application as Fig. \ref{fig_hermite} demonstrates.

A research gap is identified that remains open, which provides the motivation for this article. Methods are desired that (i) hierarchically reduce complexity from low-level tracking to high-level planning for computational efficiency, while (ii) maintaining trajectory shaping ability, in particular, not just at control points but along the entire trajectory, and (iii) reduce the number of hyperparameters to a minimum. The proposal of such a method is the contribution of this article.

Three ideas are combined: (i) high-level path planning based on the lower-dimensional Dubins airplane model \cite{mclain2014implementing,ambrosino2006algorithms}, (ii) 
a spatial modeling approach \cite{spedicato2017minimum,arrizabalaga2022towards,plessen2026simple, chan2025near}, and (iii) a layered approach where optimization (linear programming) is used for reference planning before its output is synchronized with an algebraic controller for reference tracking \cite{lee2010geometric,lopez2023pid}.

The remaining paper is organized as follows: problem formulation, solution, numerical examples, outlook, and the conclusion are described in Sect. \ref{sec_probformul}-\ref{sec_conclusion}.

\section{Problem Formulation\label{sec_probformul}}

The addressed problem can be formulated as follows: \emph{Given an initial path, that may consist of a sparse set of waypoints with larger inter-waypoint spacing and abrupt heading changes, and which in the following is for brevity referred to as an 'edgy' reference path, a control method is sought that lets a quadrotor (i) closely track the given reference, subject to (ii) lateral constraints along the path.} An application is, e.g., the tracking of reference paths, where areas next to the reference have to be avoided (obstacle areas). See Fig. \ref{fig_problvisualization} for problem visualization.

Two variations are considered. First, a smooth reference generation step is required only once offline, before this reference is then tracked online by a low-level tracking controller. Second, both the smooth reference generation and the low-level tracking step are carried out online in closed-loop. A smooth reference is planned over a limited spatial planning horizon ahead. The methods for these two variations shall in the following be referred to as methods $\textsf{M}_1$ and $\textsf{M}_2$.

\section{Problem Solution\label{sec_probsoln}}

For the problem solution it is differentiated between a high-level reference smoothing step and a low-level reference tracking step. These two steps and its combination are presented in the following three sections.

\subsection{High-level reference smoothing\label{subsec_highlevel}}

For high-level reference smoothing in 3D the method from \cite{plessen2026simple} is used. It is derived based on a Dubins airplane model with 4 states. These are the 3 position states in the inertial world frame, $(x,y,z)\in\mathbb{R}^3$, and yaw (heading) angle $\psi\in[0,2\pi]$. The equations of motion of the Dubins airplane model according to \cite{mclain2014implementing} are,
\begin{align}
\begin{bmatrix} \dot{x}\\ \dot{y} \\ \dot{z} \\ \dot{\psi} \end{bmatrix} &= \begin{bmatrix} v \cos(\gamma) \cos(\psi) \\ v \cos(\gamma) \sin(\psi) \\ v \sin(\gamma) \\  \frac{g}{v}\tan(\phi) \end{bmatrix},\label{eq_dubins}
\end{align}
with variables airspeed $v$, flight-path angle $\gamma$, roll angle $\phi$, and $g$ the acceleration due to gravity. Note that $z$-dynamics can be decoupled. Then, a spatial transformation \cite{plessen2026simple} permits to formulate a linear program (LP),
\begin{subequations}
\label{eq_LP2}
\begin{align}
\min\limits_{ \{ \phi_k \}_{k=0}^{N-1}} &\ \  \sum\nolimits_{k=1}^N |e_{y,k}- e_{y,k}^\text{ref} | \label{eq_LP2_objFcn}\\
\mathrm{s.t.} &\  \ e_{y,k} - e_{y,k}^\text{ref} \leq 0, \ k = 1,\dots,N, \label{eq_LP2_eymax_cstrts}\\
&\ \ \phi_\text{min} \leq  \phi_k \leq \phi_\text{max},  \ k = 0,\dots,N-1,\label{eq_LP2_delta_cstrts}
\end{align}
\end{subequations}
with $D_{s,k}=s_{j+k}-s_k,\forall k = 0,\dots,N-2$, and $s_k$  defining the spatial coordinate along the reference trajectory at discretization points. The lateral error, $e_y(s_k)$, is abbreviated by $e_{y,k}$, and is a nonlinear function of curvature along the reference path after the spatial transformation. By designing 
\eqref{eq_LP2_eymax_cstrts}, obstacle avoidance and other path shaping constraints can be enforced.

Several details are mentioned. First, in the original formulation of \eqref{eq_LP2_objFcn} in \cite{plessen2026simple}, also control rate constraints are included. These are here dismissed to obtain a minimum-size LP, while still maintaining shaping functionality. Second, \eqref{eq_LP2} is a small LP. After introduction of $N$ surrogate variable and after constraint softening of \eqref{eq_LP2_eymax_cstrts} by introduction of a scalar non-negative slack variable for feasibility guarantee, there are $n_p=2N$ state constraints and $n_u=2N+1$ scalar real-valued optimization variables. For perspective, for a discretization spacing of 1m and a planning horizon of 10m ahead there are only 21 optimization variables.

\begin{figure}[htbp]
\centering
%
\begin{subfigure}[b]{0.239\textwidth}
\centering
\includegraphics[width=1.0\textwidth]{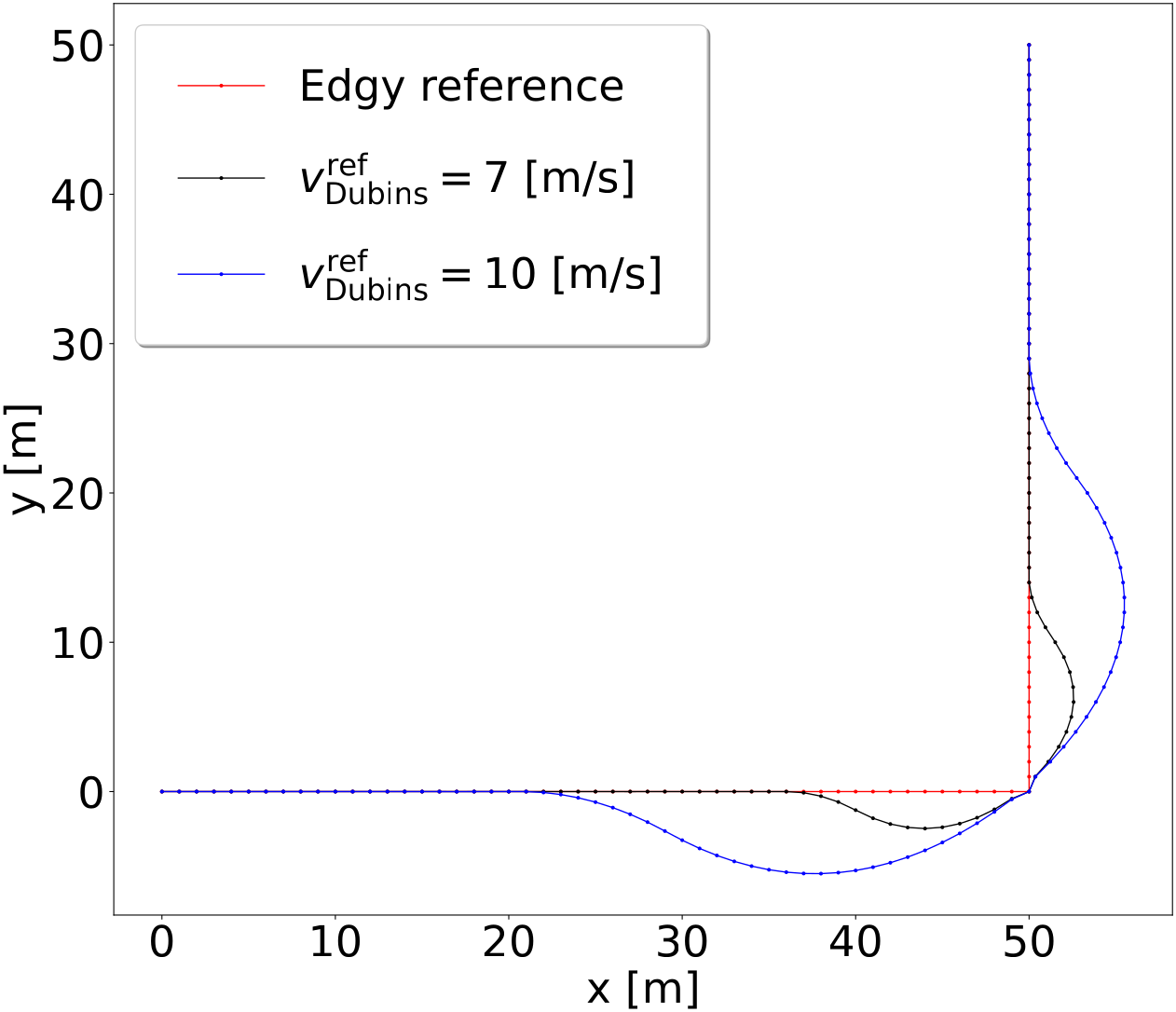}
\caption{Without any smoothing step. \vspace*{0.1cm}}
\label{fig_nosmoothingstep}
\end{subfigure}
\begin{subfigure}[b]{0.239\textwidth}
\centering
\includegraphics[width=1.0\textwidth]{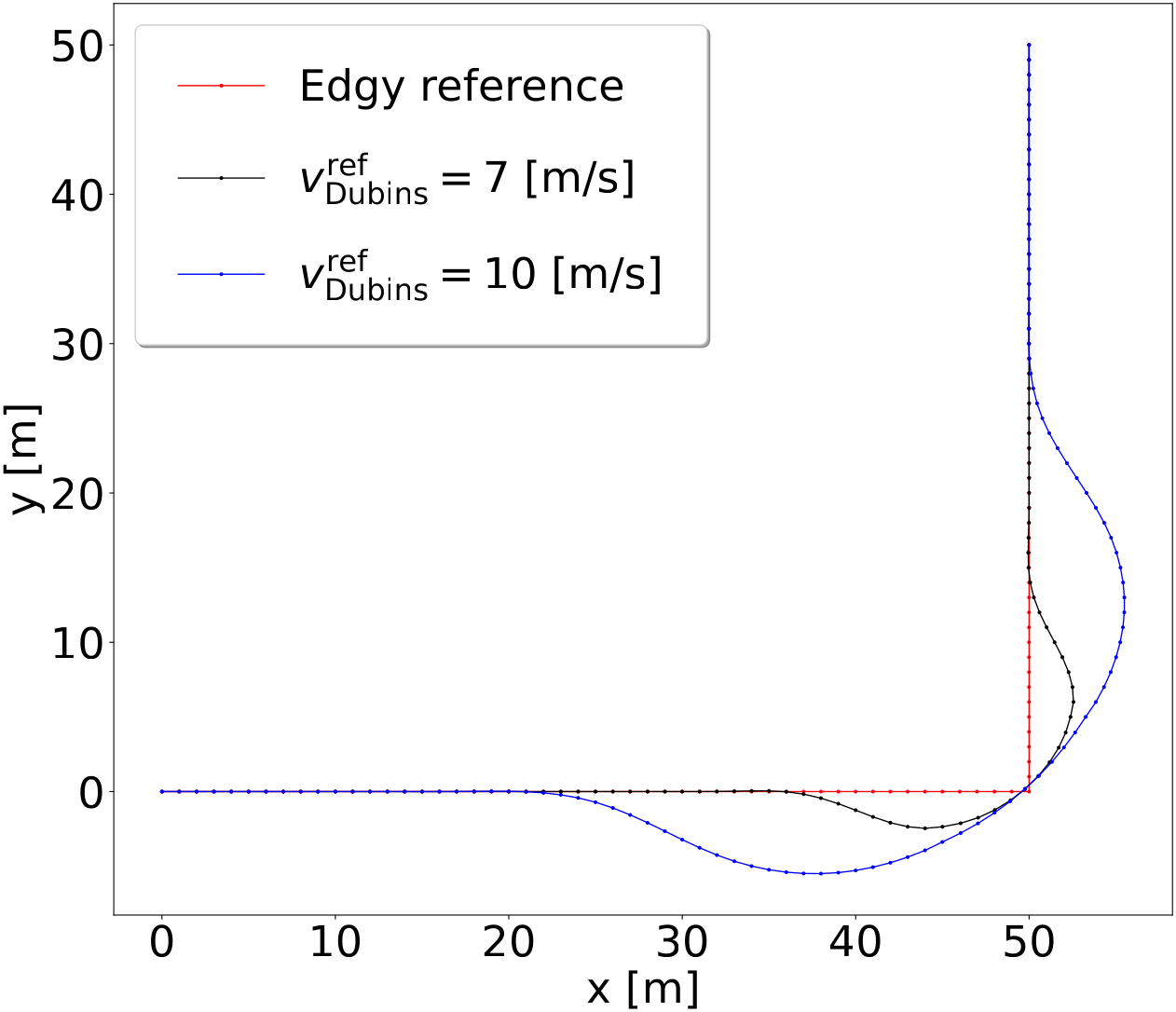}
\caption{Smoothing step: SG-filter \cite{schafer2011savitzky}.\vspace*{0.1cm}}
\label{fig_SGfilter}
\end{subfigure}
%
\begin{subfigure}[b]{0.239\textwidth}
\centering
\includegraphics[width=1.0\textwidth]{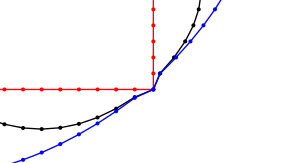}
\caption{Zoom-in. \vspace*{0.1cm}}
\end{subfigure}
\begin{subfigure}[b]{0.239\textwidth}
\centering
\includegraphics[width=1.0\textwidth]{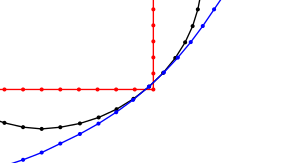}
\caption{Zoom-in. \vspace*{0.1cm}}
\label{fig_SG_zoomin}
\end{subfigure}
%
\caption{Illustration of an implementation detail. For improved smoothing a filtering step, for example, based on the SG-filter (\cite{savitzky1964analytical,schafer2011savitzky}) or EMA-filter, is applied to the trajectory resulting from the solution of LP \eqref{eq_LP2}. For algebraic tracking controllers smooth references are desirable \cite{lee2010geometric}.}
\label{fig_smoothingstep}
\end{figure}

After the solution of \eqref{eq_LP2}, a filtering step is applied for detail-smoothing. See Fig. \ref{fig_smoothingstep} for illustration and motivation. The Savitzky-Golay (SG) filter (\cite{savitzky1964analytical,schafer2011savitzky}) and the algebraic classical Exponential Moving Average (EMA) filter (i.e., $x_i^\text{EMA}=\alpha x_i + (1-\alpha)x_{i-1}^\text{EMA}$, $x_0^\text{EMA}=x_0$ and, e.g., $\alpha=0.5$ for a sequence $\{x_i\}$) are considered, and applied channel-wise to the sequence of $x$, $y$ and $z$-coordinates, respectively. The improved smoothing comes at the cost of minuscule corner-cutting for edgy initial references, see Fig. \ref{fig_SG_zoomin}.

Several comments are made. First, based on the yaw equation a minimum turning radius can be derived. The radius of curvature can be related to speed as, $v=\rho_s \dot{\psi}=\rho_s \frac{g \tan(\phi)}{v}$. This relation can be used for trajectory shaping. Trajectories with desired larger or smaller radius of curvature can be shaped by varying reference speed of the Dubins airplane model as a hyperparameter. This is illustrated in Fig. \ref{fig_nosmoothingstep} for two velocities.

Second, once a trajectory has been shaped, an upper rule-of-thumb-like approximated speed bound can be derived for another machine (in our case a quadrotor) for flying along this trajectory, 
\begin{equation}
v_\text{max}^\text{roll}(s)= \sqrt{ \rho_s(s) g \tan(\phi_\text{max}) },
\end{equation}
where $\phi_\text{max}$ is a roll angle limit, and $s$ represents the spatial coordinate along the trajectory of length $L$, i.e., $s\in[0,L]$.

Third, from Newton's second law of motion, and assuming simplifiedly constant deceleration during maximal braking, a second velocity bound can be derived. This is the braking profile for the transition from the current position towards a goal hovering position, $q_\text{g}=(x_q,y_q,z_g)$,
\begin{equation}
v_\text{max}^\text{brake}(p(s),q_\text{g}) = \sqrt{ 2 a_\text{max}^\text{brake} d_s },\label{eq_amaxbrake}
\end{equation}
where $a_\text{max}^\text{brake} = g\sqrt{ (T_\text{max}/(mg) )^2 -1 }$, and $d_s=\sqrt{(x(s)-x_g)^2 + (y(s)-y_g)^2 + (z(s)-z_g)^2}$, and $T_\text{max}$ denotes the maximal thrust force. The maximal deceleration for a quadrotor is derived from two force balances, $mg=T\cos(\theta)$ which yields $\theta_\text{max}\leq\arccos(mg/T_\text{max})$, and $ma^\text{brake}=-T\sin(\theta)$ which yields \eqref{eq_amaxbrake} when combining the two balances with the identity $\tan(\arccos(x))=\sqrt{(1-x^2)}/x$.

Thus, once a trajectory has been shaped, the velocity profile along this trajectory is assigned according to
\begin{equation}
v^\text{ref}(s)=\text{min}\left( v_\text{cruise}(s), v_\text{max}^\text{roll}(s) , v_\text{max}^\text{brake}(p(s),q_\text{g})\right),\label{eq_vrefs}
\end{equation}
where $v_\text{cruise}(s)$ denotes the reference cruise velocity, and where the goal (hovering) position is selected e.g. as the last position of the reference path. Based on \eqref{eq_vrefs} and a low-level control sampling time $T_s$, a non-uniform spatial grid can be interpolated along the path, which is then to be tracked by the low-level controller. The importance of a suitable velocity profile along the reference path is highlighted, especially for low-level tracking controllers with feedforward terms as outlined below in Sect. \ref{subsec_lowlevel}. Two further comments are made. First, \eqref{eq_vrefs} is spatially defined, which makes the method \eqref{eq_LP2} a natural fit, which is also spatially defined and permits to formulate spatial constraints in the LP. Second, velocity bounds in \eqref{eq_vrefs} can be refined. For example, the longitudinal force balance could be extended to account for classical aerodynamic drag forces, $ma^\text{brake}=-T\sin(\theta)-\frac{1}{2}\rho_\text{air}c_dAv^2$.

To summarize, the output of the high-level reference smoothing (HLRS) step is a trajectory, $\{x(s),y(s),z(s),\psi(s)\},~\forall s\in[0,L]$, that resulted from (i) a decoupling step, a LP-solution \eqref{eq_LP2} according to \cite{plessen2026simple} to smooth an initial (possibly edgy) reference path and to account for spatial constraints, (ii) a subsequent filtering step (SG \cite{savitzky1964analytical} or EMA), and (iii) is defined along non-uniform inter-point spacing to account for a velocity profile \eqref{eq_vrefs} along the trajectory.

\subsection{Low-level reference tracking\label{subsec_lowlevel}}

\begin{figure}
\centering
\includegraphics[width=0.75\linewidth]{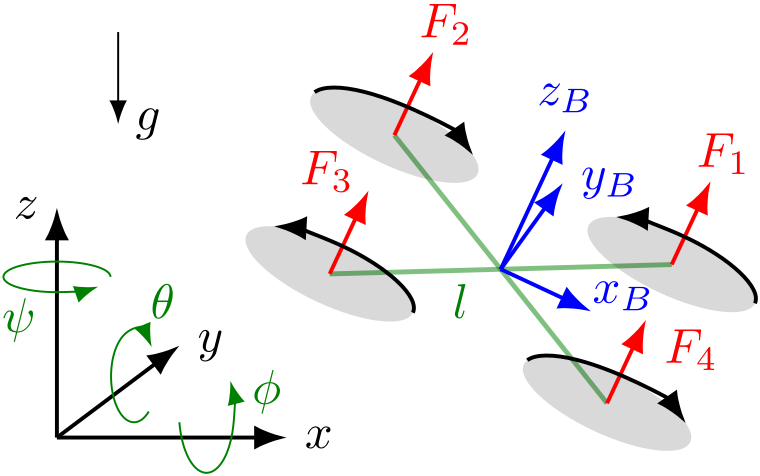}
\caption{Illustration of the quadrotor model with the world and quadrotor's body frame convention. The acceleration due to gravity is denoted by $g$. The distance (arm length) between the quadrotor's center-of-gravity and rotors is $l$. The body angular rates, $\omega_{\phi,B}$, $\omega_{\theta,B}$, $\omega_{\psi,B}$, which are not displayed for clarity, are about the $x_B$, $y_B$ and $z_B$-axes, respectively.}
\label{fig_quadrotorcoordsys}
\end{figure}

While the method above for reference smoothing was based on the  Dubins airplane model with 4 states for complexity reduction, for low-level closed-loop tracking the full 12-states quadrotor model is used. See Fig. \ref{fig_quadrotorcoordsys} for visualization. Abbreviating $c_j=\cos(j)$ for $j\in\{\phi,\theta,\psi\}$ and similarly $s_j=\sin(j)$ and $t_j=\tan(j)$, the equations of motions of the quadrotor model are,
\begin{align}
\begin{bmatrix} \dot{x}\\ \dot{y} \\ \dot{z} \\ \dot{v}_x \\ \dot{v}_y \\ \dot{v}_z \\ \dot{\phi} \\ \dot{\theta} \\ \dot{\psi} \\ \dot{\omega}_{\phi,B} \\ \dot{\omega}_{\theta,B} \\ \dot{\omega}_{\psi,B} \end{bmatrix} &= \begin{bmatrix} v_x  \\ v_y  \\ v_z \\ (s_\psi s_\phi + c_\psi s_\theta c_\phi) \frac{(T_1+T_2+T_3+T_4)}{m} \\ (-c_\psi s_\phi + s_\psi s_\theta c_\phi) \frac{(T_1+T_2+T_3+T_4)}{m} \\ -g + c_\theta c_\phi \frac{(T_1+T_2+T_3+T_4)}{m} \\ \omega_{\phi,B} + s_\phi t_\theta \omega_{\theta,B} + c_\phi t_\theta \omega_{\psi,B} \\ c_\phi \omega_{\theta,B} - s_\phi \omega_{\psi,B} \\  \frac{s_\phi}{c_\theta} \omega_{\theta,B} + \frac{c_\phi}{c_\theta} \omega_{\psi,B} \\ \frac{l (T_1+T_2-T_3-T_4) }{\sqrt{2}J_{xx}} + \frac{J_{yy}-J_{zz}}{J_{xx}}\omega_{\theta,B}\omega_{\psi,B} \\ \frac{l (-T_1+T_2+T_3-T_4) }{\sqrt{2}J_{yy}}  + \frac{J_{zz}-J_{xx}}{J_{yy}}\omega_{\phi,B}\omega_{\psi,B} \\ \frac{c_\tau (T_1-T_2+T_3-T_4)}{\sqrt{2}J_{zz}}  + \frac{J_{xx}-J_{yy}}{J_{zz}}\omega_{\phi,B}\omega_{\theta,B} \end{bmatrix},\label{eq_12quadrotor}
\end{align}
which are stated explicitly for contrasting with \eqref{eq_dubins}. Quadrotor parameters used in numerical experiments are summarized in Table \ref{tab_paramquadrotor}. The 4 control variables are the 4 thrust forces $T_i,~\forall i=1,\dots,4$.

\begin{table}
\centering
\caption{The quadrotor's parameters used in numerical experiments.\label{tab_paramquadrotor}}
 \def\arraystretch{1.0}
 \begin{tabular}{|l|l|l|l|l|}
 \hline
Param. & $m$ (kg) & $l$ (m)  & $c_\tau$ (-) & $[J_{xx},~J_{yy},~J_{zz}]$ (kgm$^2$) \\
\hline
Value & 0.8 & 0.15 & 0.01 & [1e-3,~1e-3,~1.7e-3] \\
\hline 
\end{tabular}
\vspace{-0.3cm}
\end{table}

For low-level reference tracking, the geometric method from \cite{lee2010geometric} is used for its theoretical guarantees. For example, it avoids the singularities of Euler angles and the ambiguities of quaternions in representing attitude. Note that the method is heavily reliant on feedforward terms, which are a function of the reference to be tracked. At each sampling time the 4 control signals are determined as a function of the current state with respect to a desired reference 3-dimensional position, velocity, yaw angle, acceleration, four $3\times3$ symmetric positive definite tuning matrices, and quadrotor parameters such as in Tab. \ref{tab_paramquadrotor}. The four tuning matrices are typically chosen to be diagonal with positive entries. The values used throughout numerical experiments are summarized in Table \ref{tab_hyperparamlee}. Note that for simplicity and robustness testing they are (i) chosen with uniform diagonal entries, (ii) and no tuning  was carried out except for the orders of magnitude (i.e., all scalar numbers were selected as powers of ten).

The desired reference is selected as the projection point, described by $s_{\hat{k}}\in[0,L]$, of the quadrotor's position onto the reference path determined from Sect. \ref{subsec_highlevel}, before desired velocities and accelerations are then extracted from the reference path directly via central differences. This step is possible because of the velocity profile generation and non-uniform inter-point spacing adapted to sampling time $T_s$ and quadrotor dynamics as described towards the end of Sect. \ref{subsec_highlevel}. A suitable, in general non-uniform, spatial interpolation grid along the smoothed reference path from Sect. \ref{subsec_highlevel} was found to be crucial for good closed-loop control performance. The reason is the dependency of the controller on feedforward terms. To demonstrate this sensitivity, two different calculation versions ($\textsf{v1}$ and $\textsf{v2}$) are compared below in numerical experiments of Sect. \ref{sec_expts} for the feedforward term relating to reference accelerations,
\begin{subequations}
\label{eq_ddpd}
\begin{align}
 a_{j}^{\text{ref},\textsf{v1}}(s_{\hat{k}}) &=   \left( v_{j}^\text{ref}(s_{\hat{k}+1})-v_{j}^\text{ref}(s_{\hat{k}}) \right)/T_s,  \label{eq_ddpd_v1}\\
a_{j}^{\text{ref},\textsf{v2}}(s_{\hat{k}})  &=   \frac{2}{T_s^2} ( j^\text{ref}(s_{\hat{k}+1})-j^\text{ref}(s_{\hat{k}}) - T_s v_{j}^\text{ref}(s_{\hat{k}}) ), \label{eq_ddpd_v2}
\end{align}
\end{subequations}
for $j\in\{x,y,z\}$, and where a given discretized reference trajectory according to Sect. \ref{subsec_highlevel} is assumed.

\begin{table}
\centering
\caption{Hyperparameters selected for the controller \cite{lee2010geometric} according to Sect. \ref{subsec_lowlevel}. The identity matrix of dimension 3 is denoted by $I_3$.\label{tab_hyperparamlee}}
 \def\arraystretch{1.0}
 \begin{tabular}{|l|l|l|l|l|}
 \hline
Param. & $K_p$ & $K_v$  & $K_R$ & $K_\Omega$  \\
\hline
Value & $10 I_3$ & $I_3$ & $0.1I_3$ & $0.01I_3$ \\
\hline 
\end{tabular}
\vspace{-0.3cm}
\end{table}


\subsection{Methods $\textsf{M}_1$ and $\textsf{M}_2$\label{subsec_M1M2}}

\begin{algorithm}
\SetKwInOut{Subfunctions}{\textbf{Subfunction}}
\SetKwInOut{Input}{\textbf{Data Input}}
\SetKwInOut{Output}{\textbf{Data Output}}
\DontPrintSemicolon
\vspace{0.15cm}
\Subfunctions{$\mathcal{F}(\cdot)$,~$\mathcal{F}_{\textsf{M}_1}^\text{HLRS}(\cdot)$,~$\mathcal{F}^\text{LLT}(\cdot)$.}
\vspace{0.15cm}\hrule\vspace{0.15cm}
%
\Input{$\{(x,y,z)\},~\tilde{p}_0,~T_s$.}
%
\vspace{0.15cm}\hrule\vspace{0.15cm}
$\mathcal{Q} \leftarrow \{(x,y,z)\}$~{\color{gray}\%Given (sparse) waypoints reference.}\;\vspace{0.15cm}
${\color{blue}\mathcal{Q}_\text{HLRS} \leftarrow \mathcal{F}_{\textsf{M}_1}^\text{HLRS}(\mathcal{Q}) }$~{\color{gray}\%High-Level Ref. Smoothing.}\;\vspace{0.15cm}
$p_0 \leftarrow \tilde{p}_0 $~{\color{gray}\%Initial quadrotor state.}\;\vspace{0.15cm}
\For{$t\in\{0,T_s,2T_s,\dots\}$}
{\vspace{0.15cm}
${\color{blue}u_t \leftarrow \mathcal{F}^\text{LLT}(p_t,\mathcal{Q}_\text{HLRS})}$~{\color{gray}\%Low-Level Tracking.}\;\vspace{0.15cm}
$p_{t+1} \leftarrow \mathcal{F}(p_t,u_t)$~{\color{gray}\%Apply control signal to system.}\;\vspace{0.15cm}
%
%
} 
\vspace{0.0cm}\hrule\vspace{0.15cm}
\Output{$ \{ p_t \}$~{\color{gray}\%Cl.-l. traj. that tracks $\mathcal{Q}_\text{HLRS}$.}}
\vspace{0.05cm}
\caption{One-time high-level reference smoothing over the entire path horizon (offline)}\label{alg_M1}
\end{algorithm}
 
The high-level reference smoothing method from Sect. \ref{subsec_highlevel} and the low-level tracking (LLT) method from Sect. \ref{subsec_lowlevel} can be combined in two ways. First, a given reference trajectory is smoothed offline, before the smoothed reference is then tracked online in closed-loop. Or, secondly, a given reference trajectory is smoothed recedingly in closed-loop directly before the low-level tracking step by solving LP \eqref{eq_LP2} over a limited spatial prediction horizon.

These two methods are referred to as $\textsf{M}_1$ and $\textsf{M}_2$ in the following, and are summarized in Alg. \ref{alg_M1} and \ref{alg_M2}, respectively. Method $\textsf{M}_1$ may be relevant, for example, for agricultural applications \cite{luna2023spiral,plessen2025path,plessen20262d} where area coverage paths are calculated once before mission start and the environment is static. In contrast method $\textsf{M}_2$ is relevant for dynamic obstacle avoidance or when waypoints may change online. 

\begin{figure}[htbp]
\centering
%
\begin{subfigure}[b]{0.32\textwidth}
\centering
\includegraphics[width=1.0\textwidth]{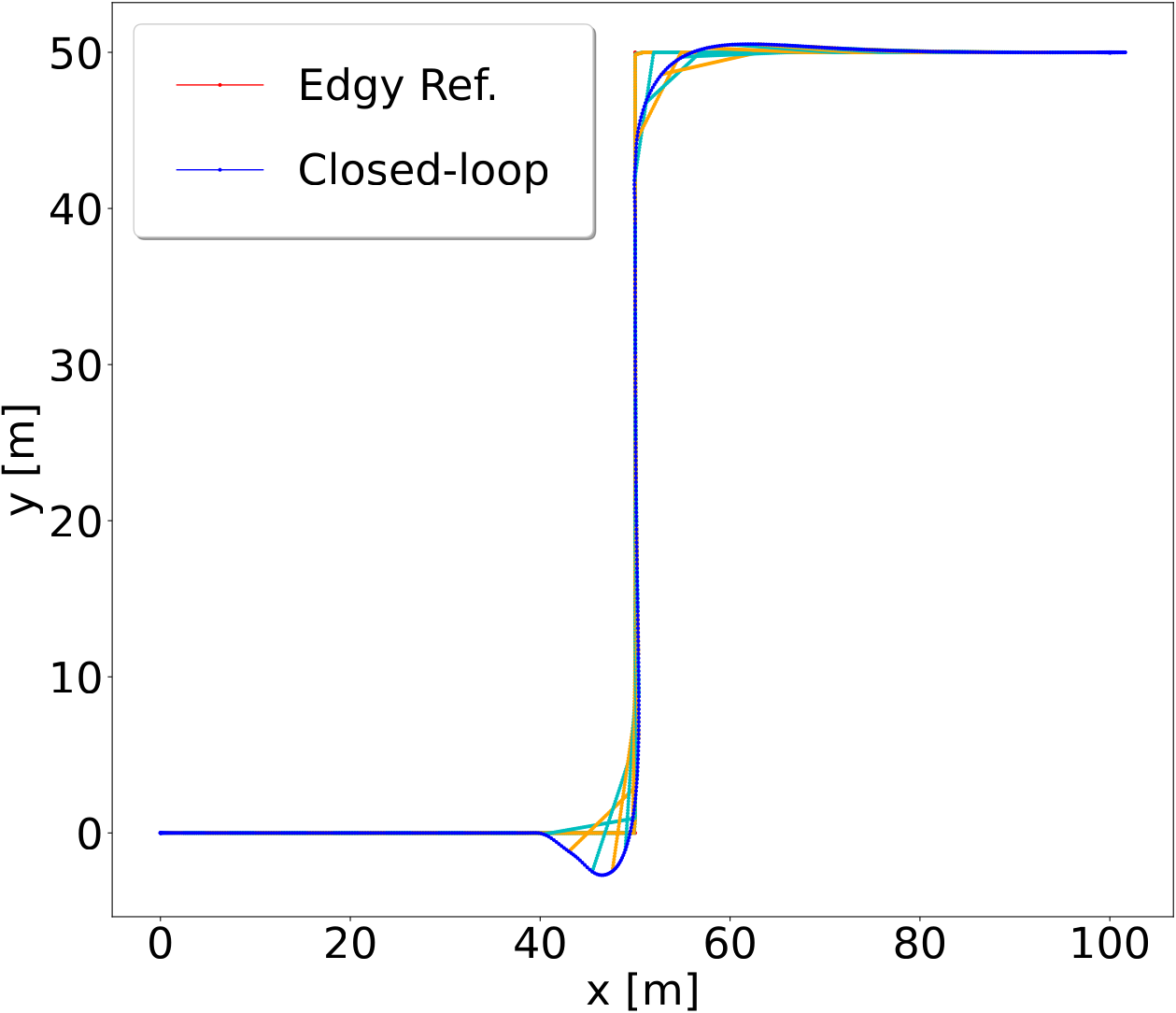}
\caption{Illustration of reference generation for M2. \vspace*{0.1cm}}
\label{fig_refgen_M2_a}
\end{subfigure}
\begin{subfigure}[b]{0.15\textwidth}
\centering
\includegraphics[width=0.93\textwidth]{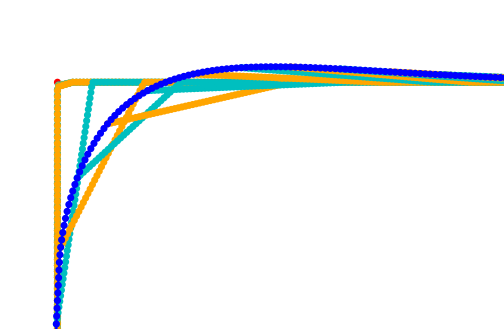} 
\caption{Zoom-in.\vspace*{0.02cm}}
\includegraphics[width=0.93\textwidth]{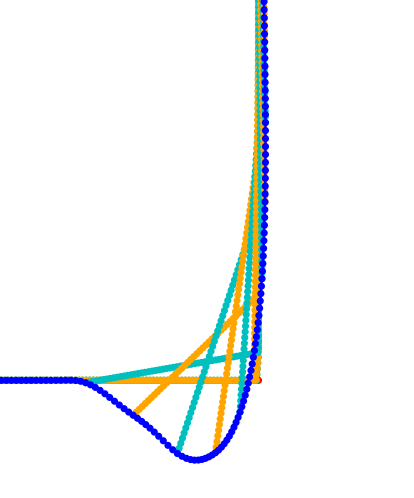}
\caption{Zoom-in.\vspace*{0.2cm}}
\label{fig_refgen_M2_c}
\end{subfigure}
%
\caption{Illustration of receding reference generation over spatial prediction horizon $H$ for $\textsf{M}_2$. The references, which are illustrated in cyan and orange and sampled only every 10th sampling time for better visualization, must connect current quadrotor position with the original edgy reference path. Each of these references is then smoothed by the method from Sect. \ref{subsec_highlevel}.}
\label{fig_refgen_M2}
\end{figure}

For $\textsf{M}_2$, Fig. \ref{fig_refgen_M2} illustrates receding reference generation over spatial prediction horizon $H$. The references must connect current quadrotor position with the original edgy reference path. This is done by (i) calculating the projection point of the current quadrotor position onto the original edgy reference, (ii) before along the original reference a spatial path segment of horizon length $H$ is interpolated, (iii) before the reference is generated by connecting the current quadrotor's position, with the point at half the spatial path segment and with the path segment onwards from that point. Each of these references is then smoothed by the method from Sect. \ref{subsec_highlevel}.

\begin{algorithm}
\SetKwInOut{Subfunctions}{\textbf{Subfunction}}
\SetKwInOut{Input}{\textbf{Data Input}}
\SetKwInOut{Output}{\textbf{Data Output}}
\DontPrintSemicolon
\vspace{0.15cm}
\Subfunctions{$\mathcal{F}(\cdot)$,~$\mathcal{F}_{\textsf{M}_2}^\text{HLRS}(\cdot)$,~$\mathcal{F}^\text{LLT}(\cdot)$.}
\vspace{0.15cm}\hrule\vspace{0.15cm}
%
\Input{$\{(x,y,z)\},~\tilde{p}_0,~T_s,~H$.}
%
\vspace{0.15cm}\hrule\vspace{0.15cm}
$\mathcal{Q} \leftarrow \{(x,y,z)\}$~{\color{gray}\%Given (sparse) waypoints reference.}\;\vspace{0.15cm}
$p_0 \leftarrow \tilde{p}_0 $~{\color{gray}\%Initial quadrotor state.}\;\vspace{0.15cm}
\For{$t\in\{0,T_s,2T_s,\dots\}$}
{\vspace{0.15cm}
${\color{blue}\mathcal{Q}_\text{HLRS} \leftarrow \mathcal{F}_{\textsf{M}_2}^\text{HLRS}(p_t,H,\mathcal{Q}) }${\color{gray}\%High-Level Reference.}\;\vspace{0.15cm}
${\color{blue}u_t \leftarrow \mathcal{F}^\text{LLT}(p_t,\mathcal{Q}_\text{HLRS})}$~{\color{gray}\%Low-Level Tracking.}\;\vspace{0.15cm}
$p_{t+1} \leftarrow \mathcal{F}(p_t,u_t)$~{\color{gray}\%Apply control signal to system.}\;\vspace{0.15cm}
%
%
} 
\vspace{0.0cm}\hrule\vspace{0.15cm}
\Output{$ \{ p_t \}$~{\color{gray}\%Cl.-loop traj. that tracks $\mathcal{Q}$.}}
\vspace{0.05cm}
\caption{Receding high-level reference smoothing over limited spatial prediction horizon $H$ (online)}\label{alg_M2}
\end{algorithm}

\section{Numerical Experiments\label{sec_expts}}

The results of 3 numerical experiments are presented, whereby each experiment is conducted for both the $\textsf{M}_1$ and $\textsf{M}_2$ methods. Each experiment consists of (i) an edgy reference path with few waypoints, large inter-waypoint spacing and $90^\circ$-heading changes, (ii) at the last waypoint a hovering state is desired, and (iii) the area left of the edgy reference path is to be avoided. The sampling time is set as $T_s=0.05$s and the dynamics \eqref{eq_12quadrotor} are integrated using the 4th-order Runge-Kutta scheme for closed-loop simulations. The reference quadrotor cruise velocity is set as $v_\text{cruise}=4$m/s.

All simulations were run on a laptop running Ubuntu 24.04 equipped with an Intel Core i9 CPU @5.50GHz×32 and 32 GB of memory. For the solution of the LPs Scipy's (cf. \cite{virtanen2020scipy}) \texttt{linprog}-solver in default settings was used.

\begin{figure}[htbp]
\centering
%
\begin{subfigure}[b]{0.239\textwidth}
\centering
\includegraphics[width=1.0\textwidth]{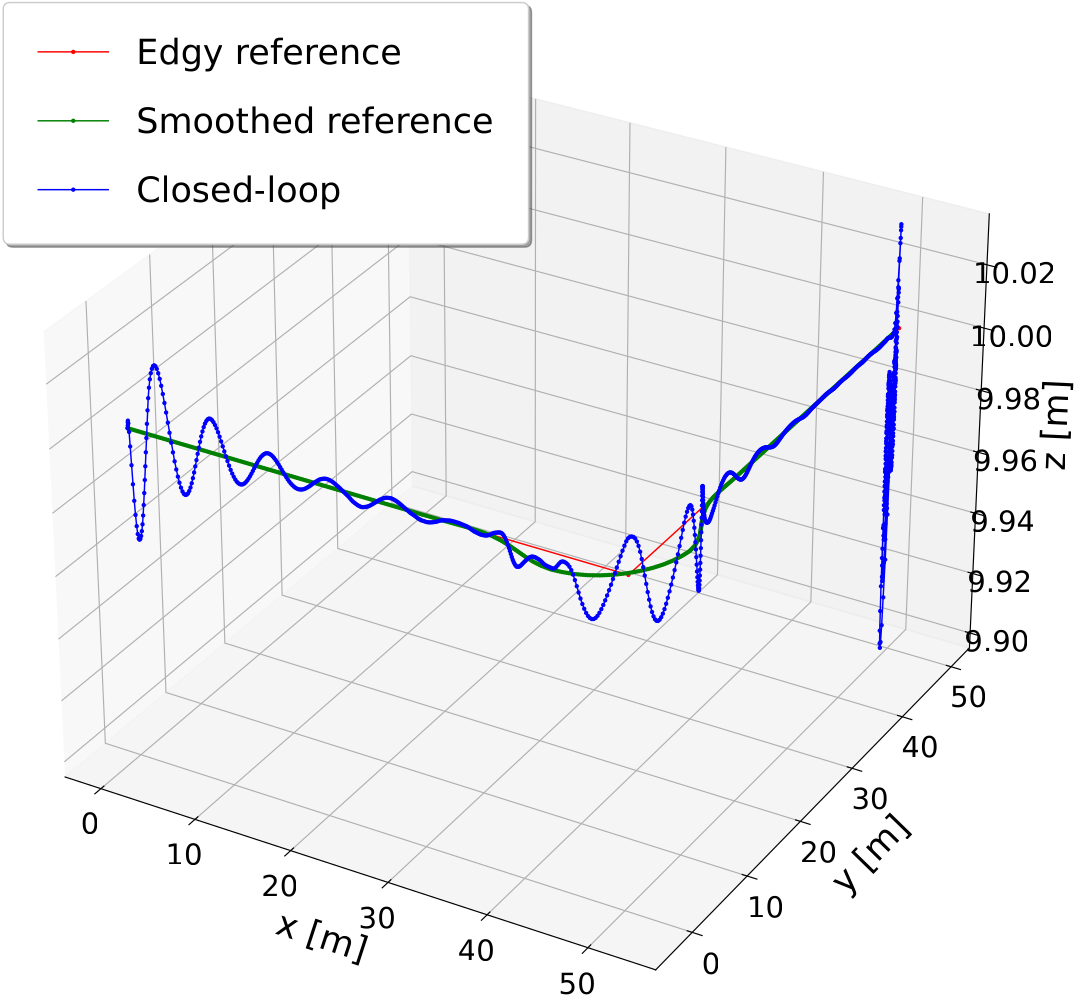}
\caption{Cascaded PID. \vspace*{0.1cm}}
\label{fig_ex1m1a}
\end{subfigure}
\begin{subfigure}[b]{0.239\textwidth}
\centering
\includegraphics[width=1.0\textwidth]{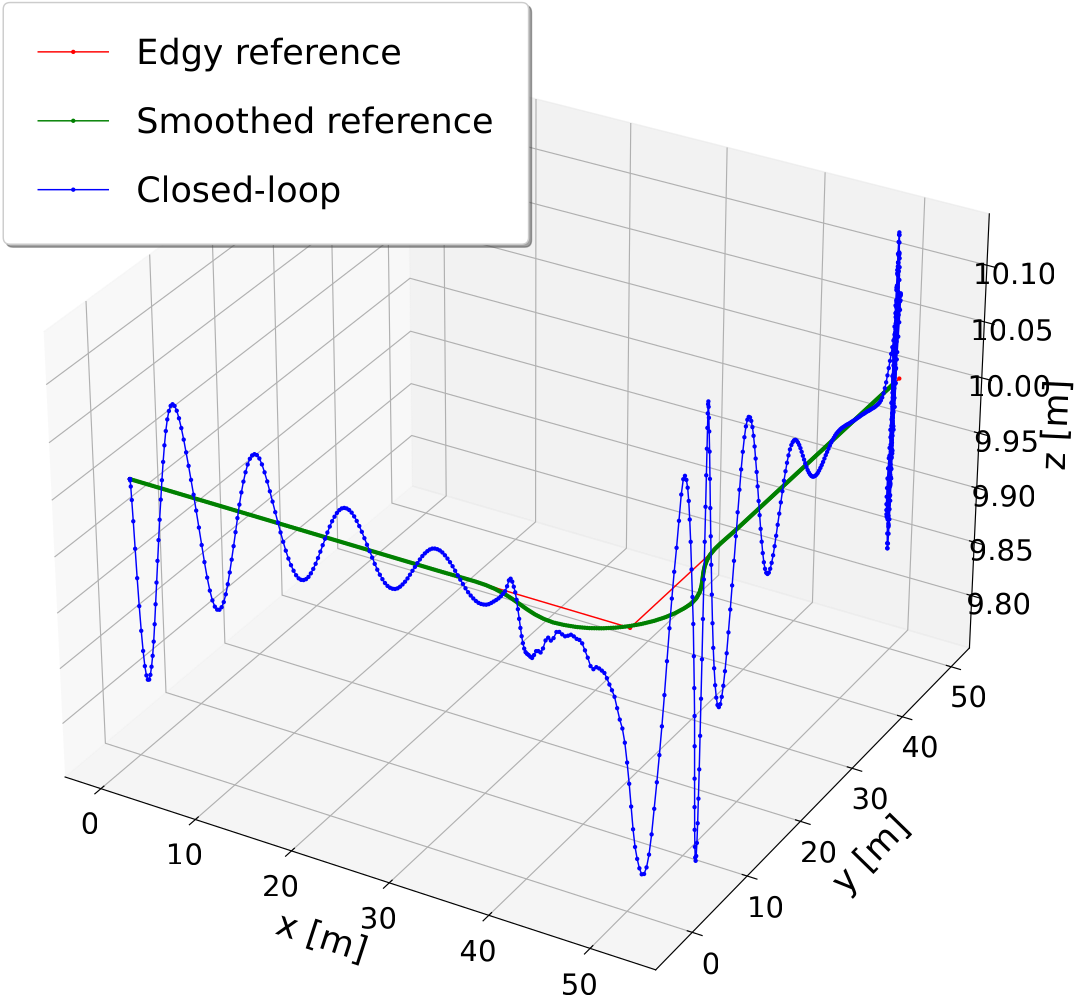}
\caption{Geometric, with Eq. \eqref{eq_ddpd_v1}.\vspace*{0.1cm}}
\label{fig_ex1m1b}
\end{subfigure}
%
\begin{subfigure}[b]{0.239\textwidth}
\centering
\includegraphics[width=1.0\textwidth]{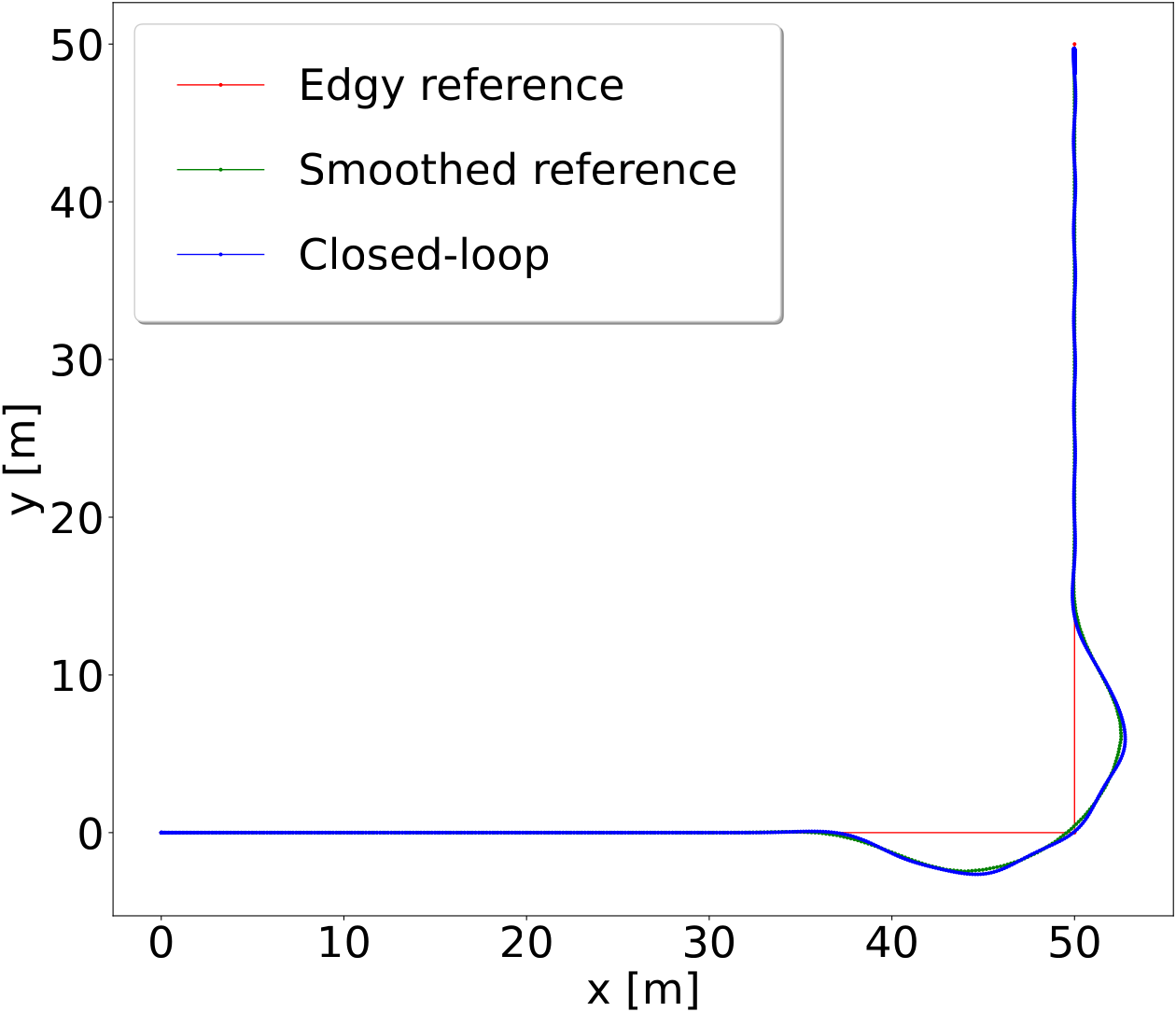}
\caption{Cascaded PID, $xy$-plane. \vspace*{0.1cm}}
\label{fig_ex1m1c}
\end{subfigure}
\begin{subfigure}[b]{0.239\textwidth}
\centering
\includegraphics[width=1.0\textwidth]{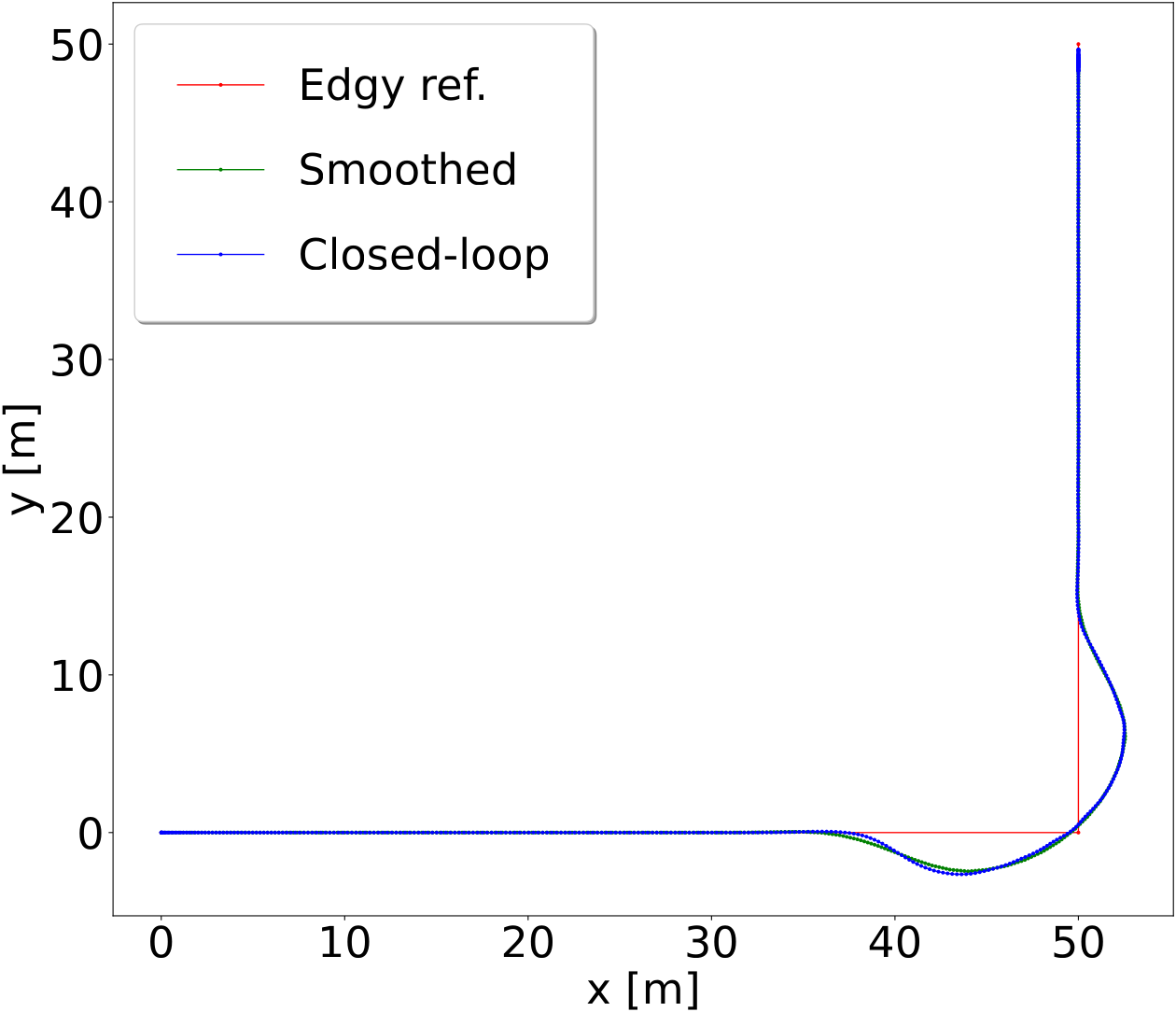}
\caption{Geometric, $xy$-plane. \vspace*{0.1cm}}
\label{fig_ex1m1d}
\end{subfigure}
%
\caption{Results for Example 1 and $\textsf{M}_1$. Comparison of 2 low-level tracking control methods: cascaded PID and the geometric tracking controller from \cite{lee2010geometric}. The z-level range is closely bounded around the reference height 10m.}
\label{fig_ex1_pid}
\end{figure}

\begin{figure}
\centering
\includegraphics[width=0.8\linewidth]{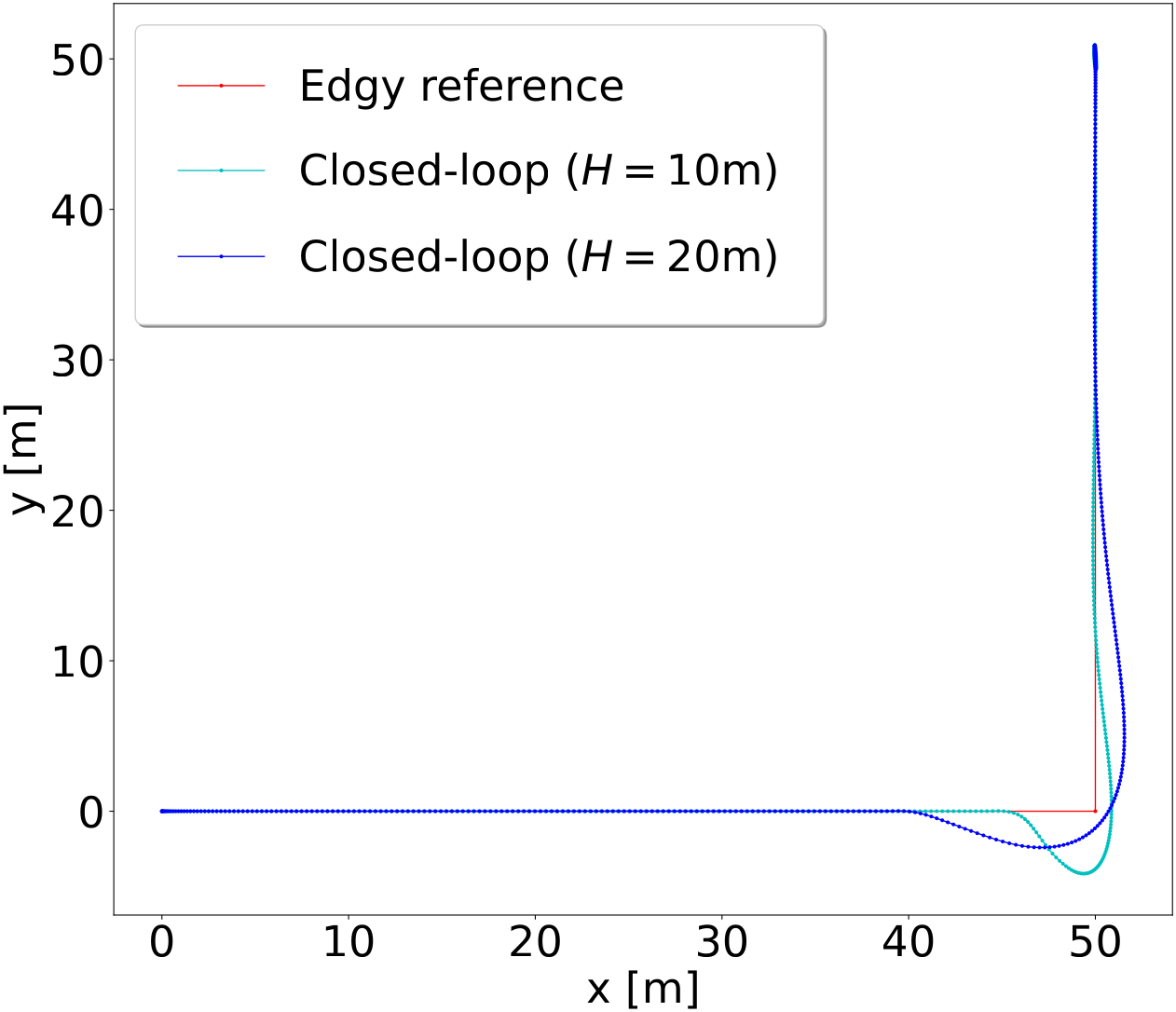}
\caption{Results for Example 1 and $\textsf{M}_2$. The influence of different spatial prediction horizons $H$ is visualized. The closed-loop tracking results are compared for $H=10$m and $H=20$m. The benefit of a larger prediction horizon is evident, however comes at larger computational cost with solve times $\bar{T}_\text{avg}^\text{LP}=0.003$s and $\bar{T}_\text{avg}^\text{LP}=0.008$s, respectively.}
\label{fig_influenceH}
\end{figure}

\begin{figure}[htbp]
\centering
%
\begin{subfigure}[b]{0.239\textwidth}
\centering
\includegraphics[width=1.0\textwidth]{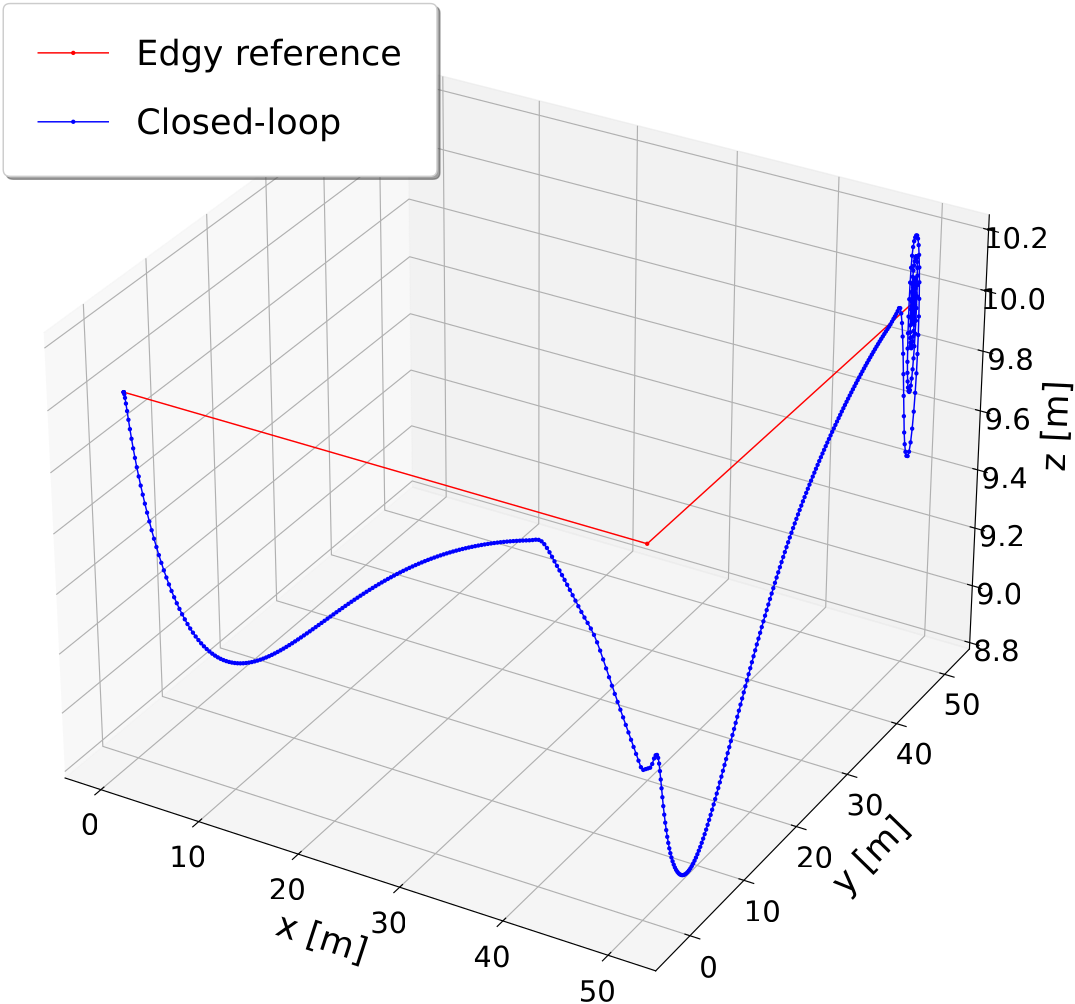}
\caption{Result in 3D. \vspace*{0.1cm}}
\label{fig_ex1m2a}
\end{subfigure}
\begin{subfigure}[b]{0.239\textwidth}
\centering
\includegraphics[width=1.0\textwidth]{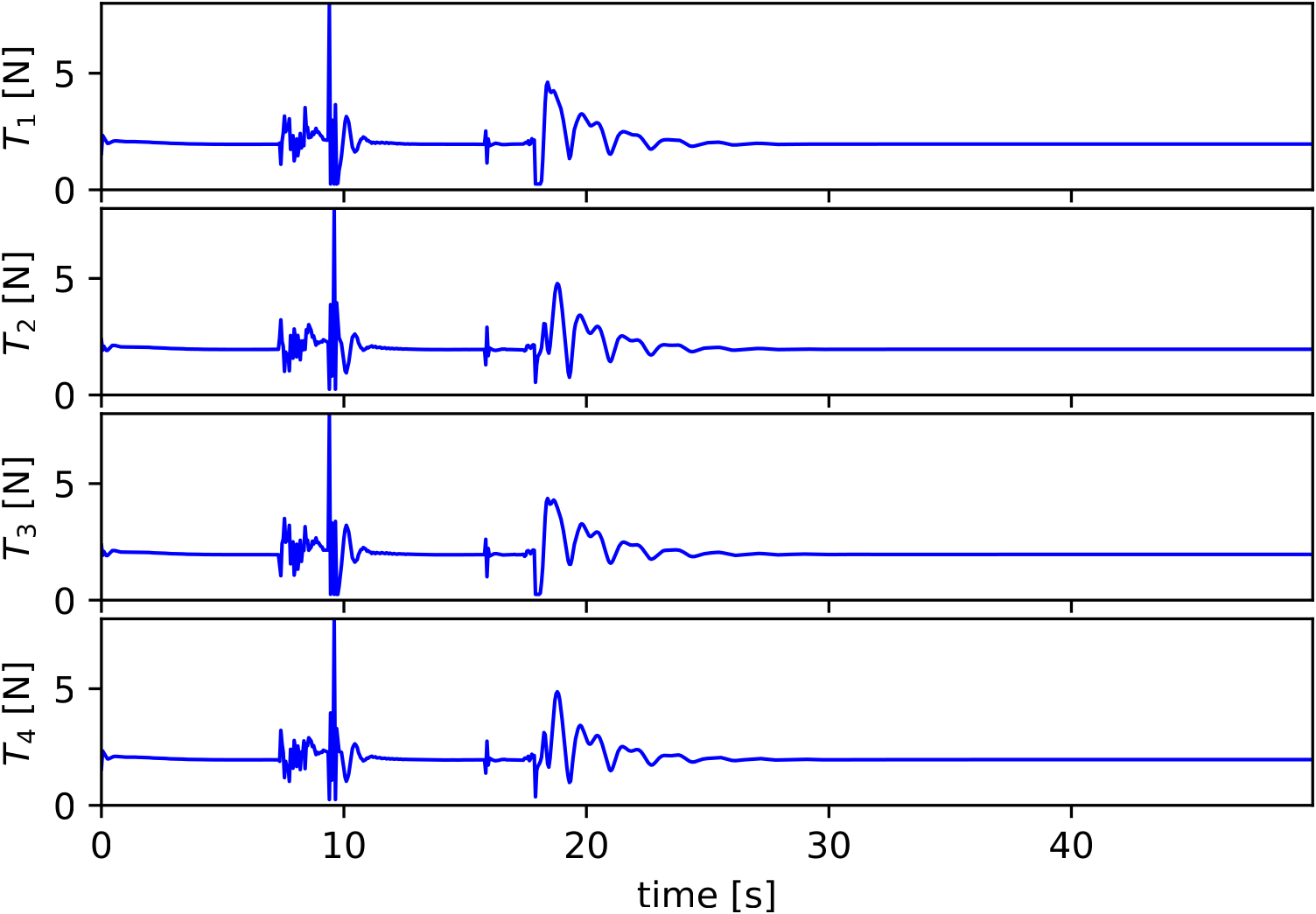}
\caption{4 control signals.\vspace*{0.1cm}}
\label{fig_ex1m2b}
\end{subfigure}
%
\begin{subfigure}[b]{0.39\textwidth}
\centering
\includegraphics[width=0.9\textwidth]{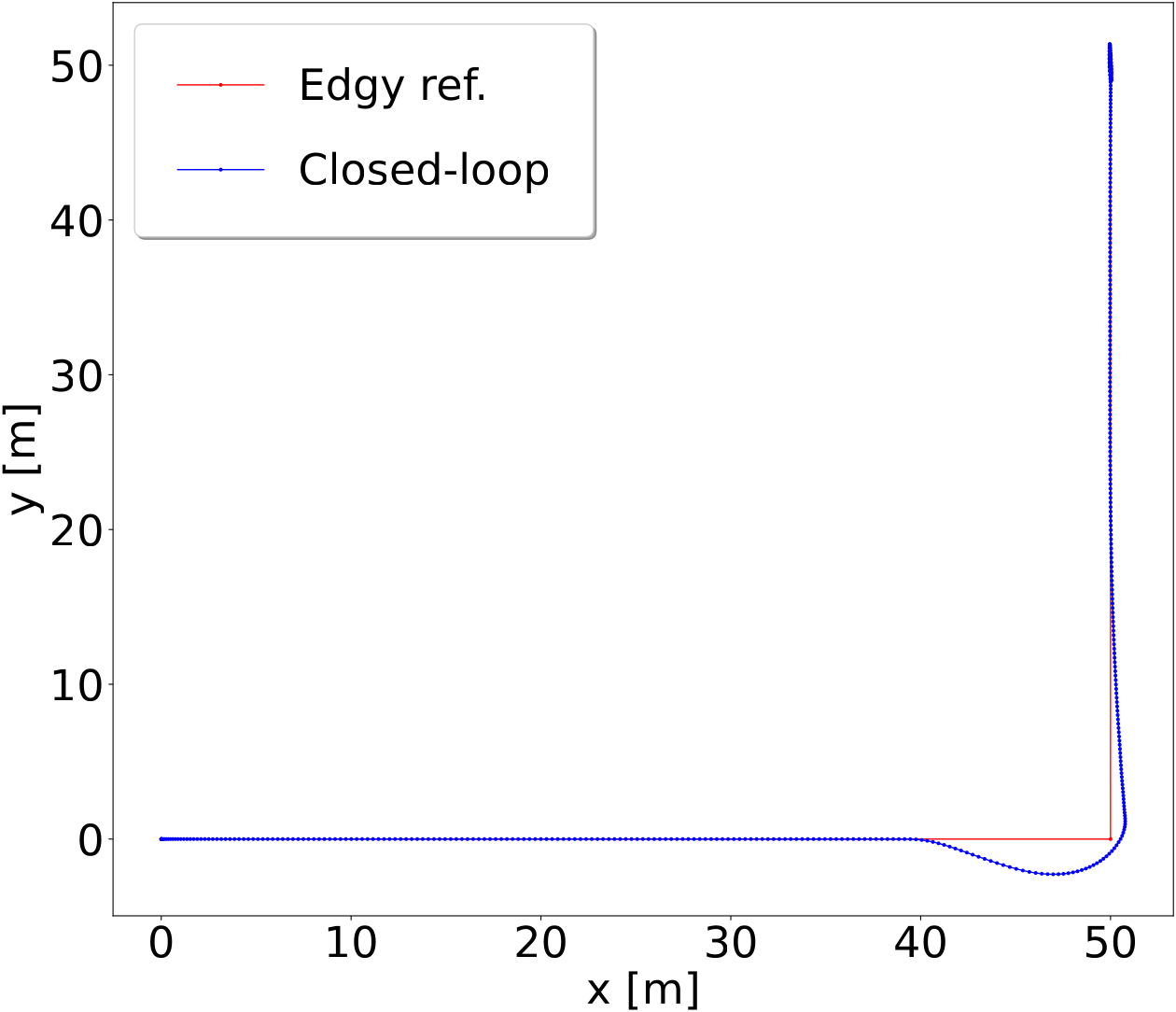}
\caption{Closed-loop tracking result in the $xy$-plane. \vspace*{0.1cm}}
\label{fig_ex1m2c}
\end{subfigure}
%
\caption{Results for Example 1 and $\textsf{M}_2$. An EMA-filtering step is applied. The reference end position, $(x,y,z)=(50,50,10)$, is reached at approximately 25s, when the quadrotor enters a hovering state at that end position.}
\label{fig_ex1m2d}
\end{figure}

\begin{figure}[htbp]
\centering
%
\begin{subfigure}[b]{0.239\textwidth}
\centering
\includegraphics[width=1.0\textwidth]{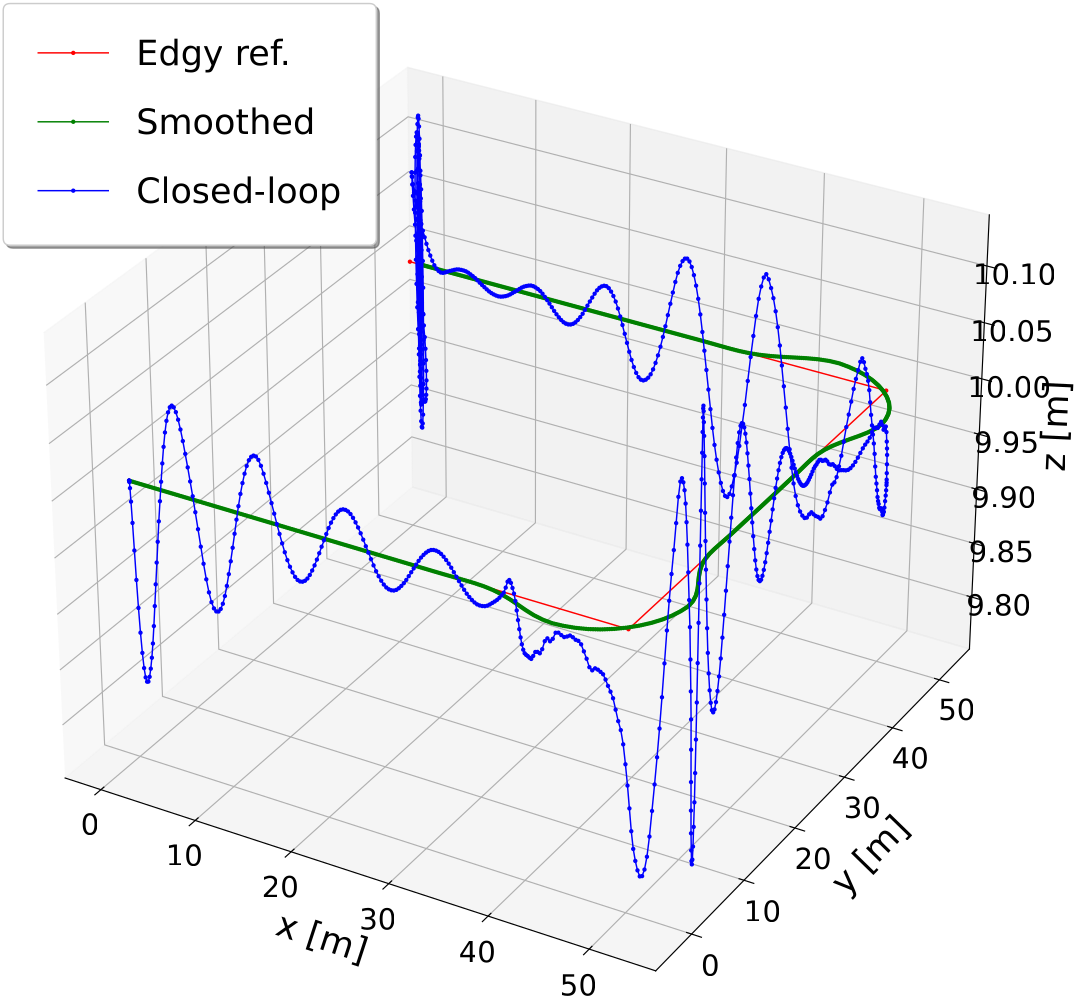}
\caption{Result in 3D, with Eq. \eqref{eq_ddpd_v1}. \vspace*{0.1cm}}
\label{fig_ex2m1a}
\end{subfigure}
\begin{subfigure}[b]{0.239\textwidth}
\centering
\includegraphics[width=1.0\textwidth]{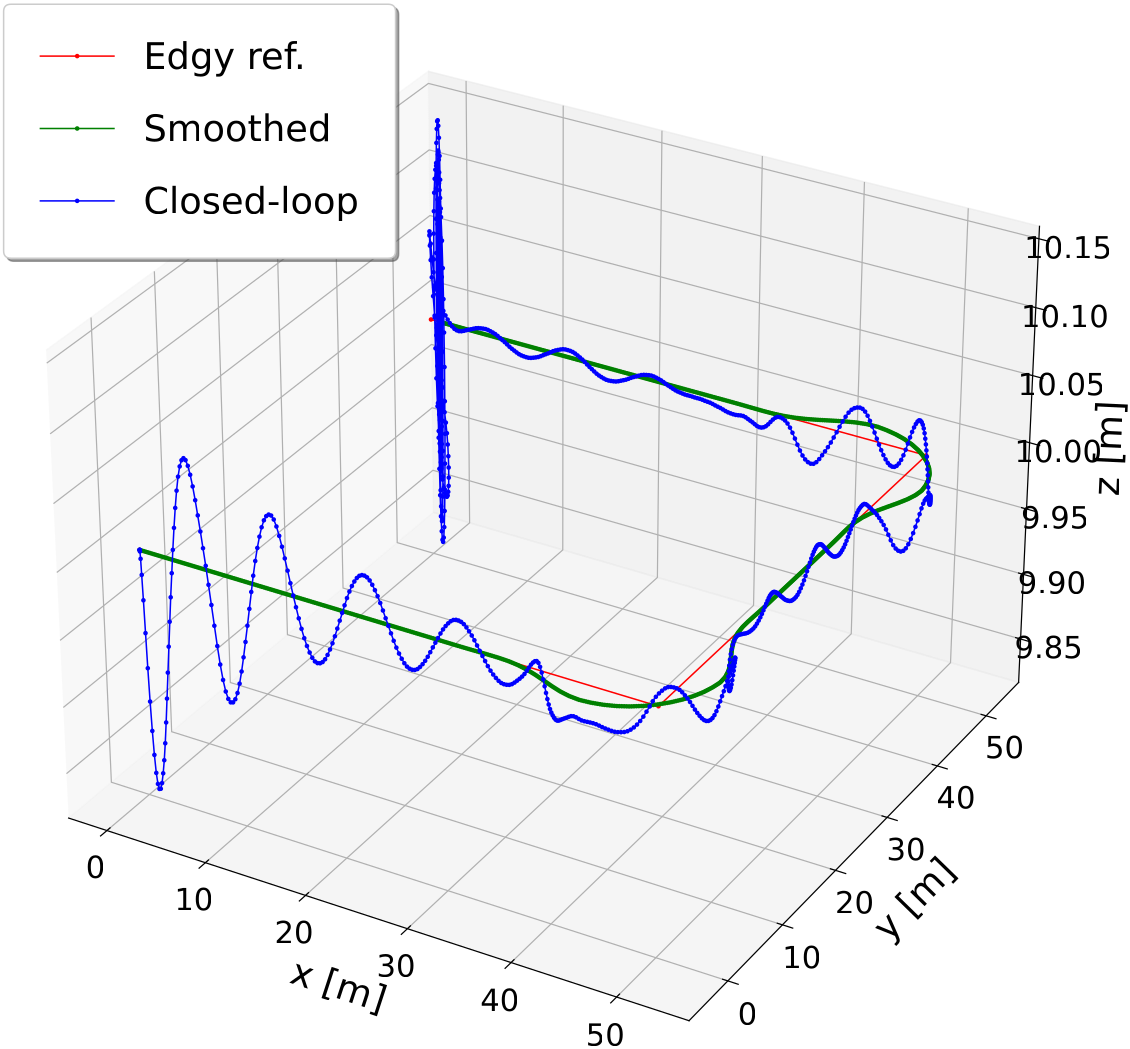}
\caption{Result in 3D, with Eq. \eqref{eq_ddpd_v2}.\vspace*{0.1cm}}
\label{fig_ex2m1b}
\end{subfigure}
%
\begin{subfigure}[b]{0.239\textwidth}
\centering
\includegraphics[width=1.0\textwidth]{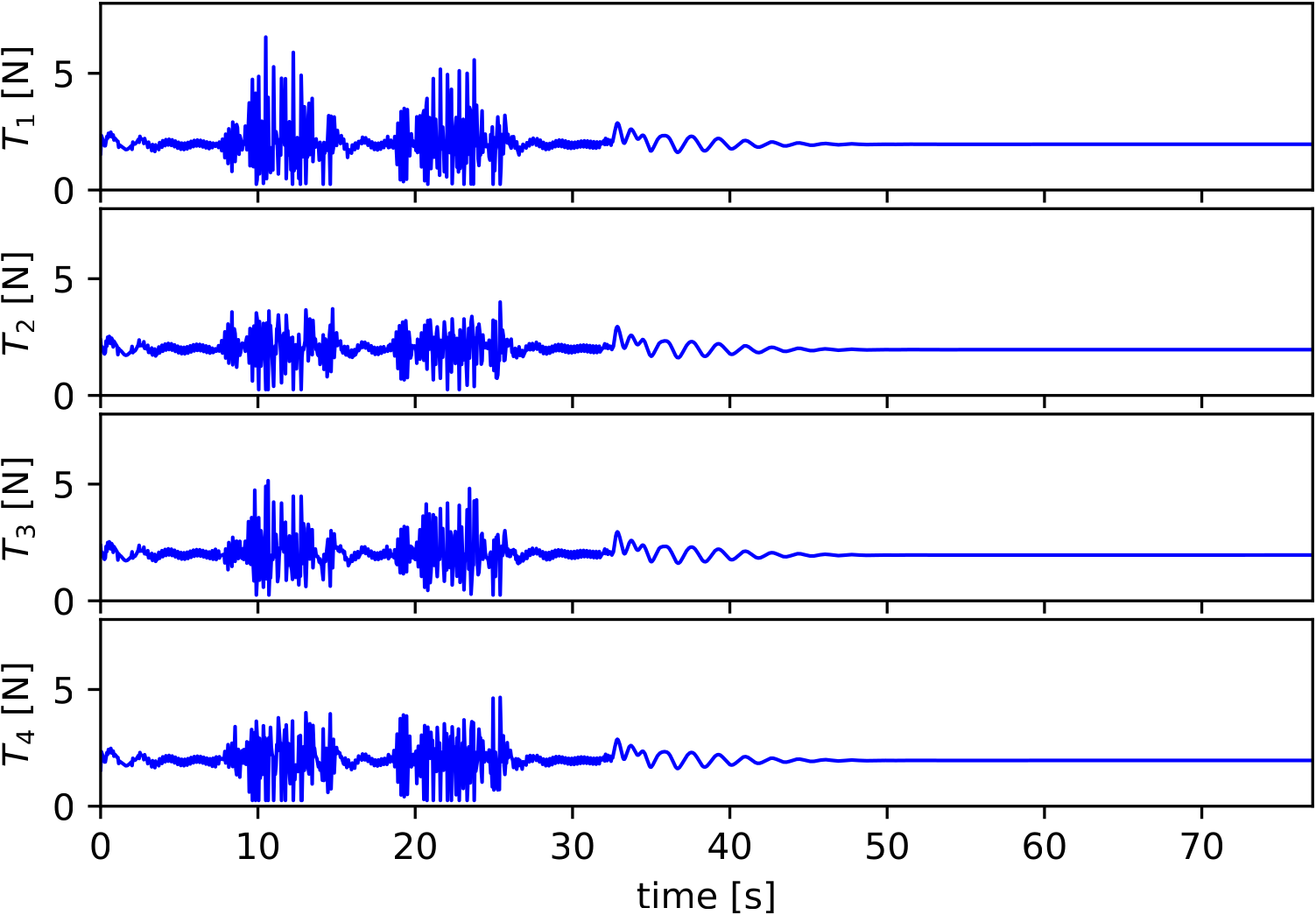}
\caption{4 control signals. \vspace*{0.1cm}}
\label{fig_ex2m1c}
\end{subfigure}
\begin{subfigure}[b]{0.239\textwidth}
\centering
\includegraphics[width=1.0\textwidth]{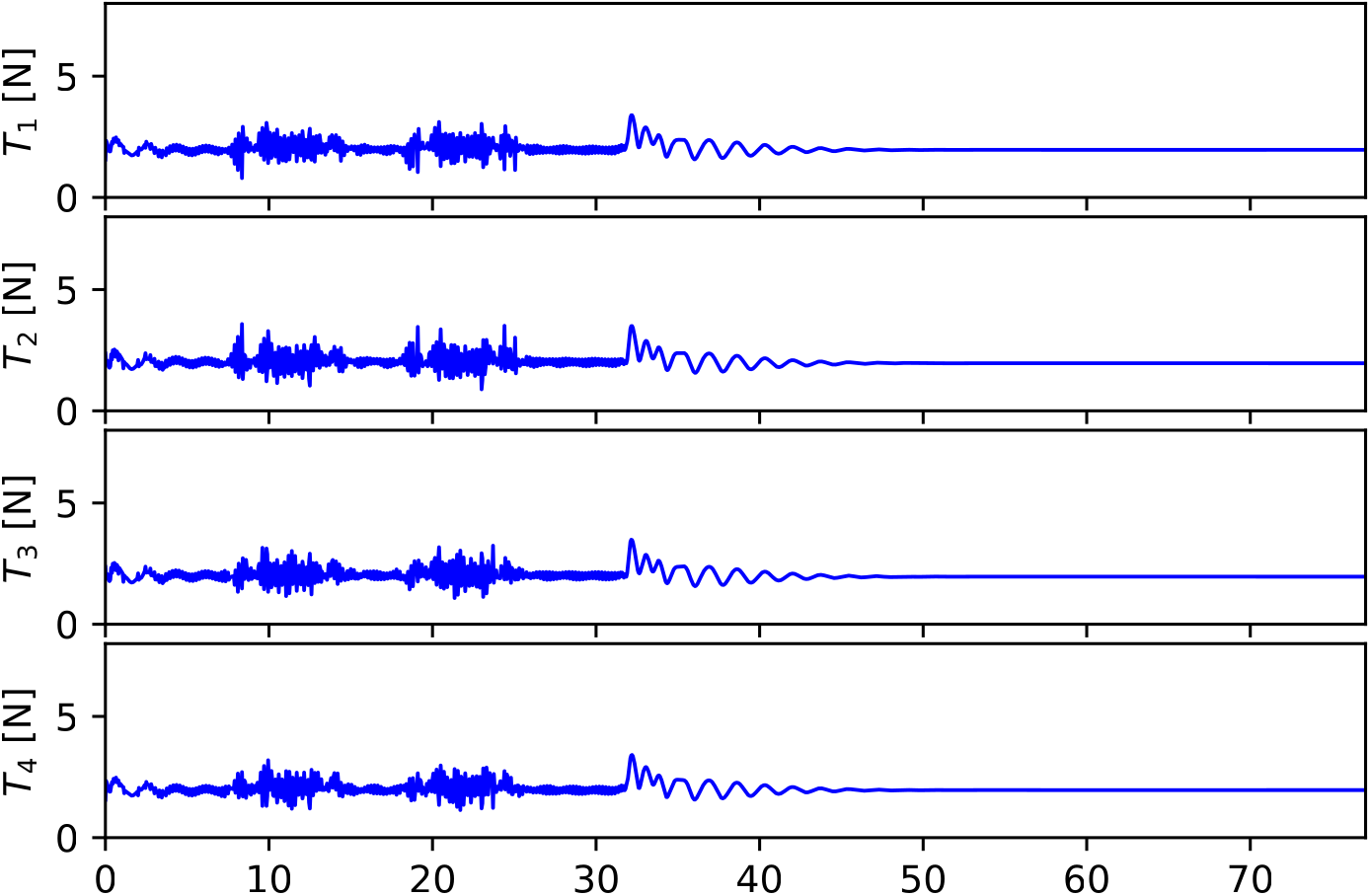}
\caption{4 control signals.\vspace*{0.1cm}}
\label{fig_ex2m1d}
\end{subfigure}
%
\begin{subfigure}[b]{0.239\textwidth}
\centering
\includegraphics[width=1.0\textwidth]{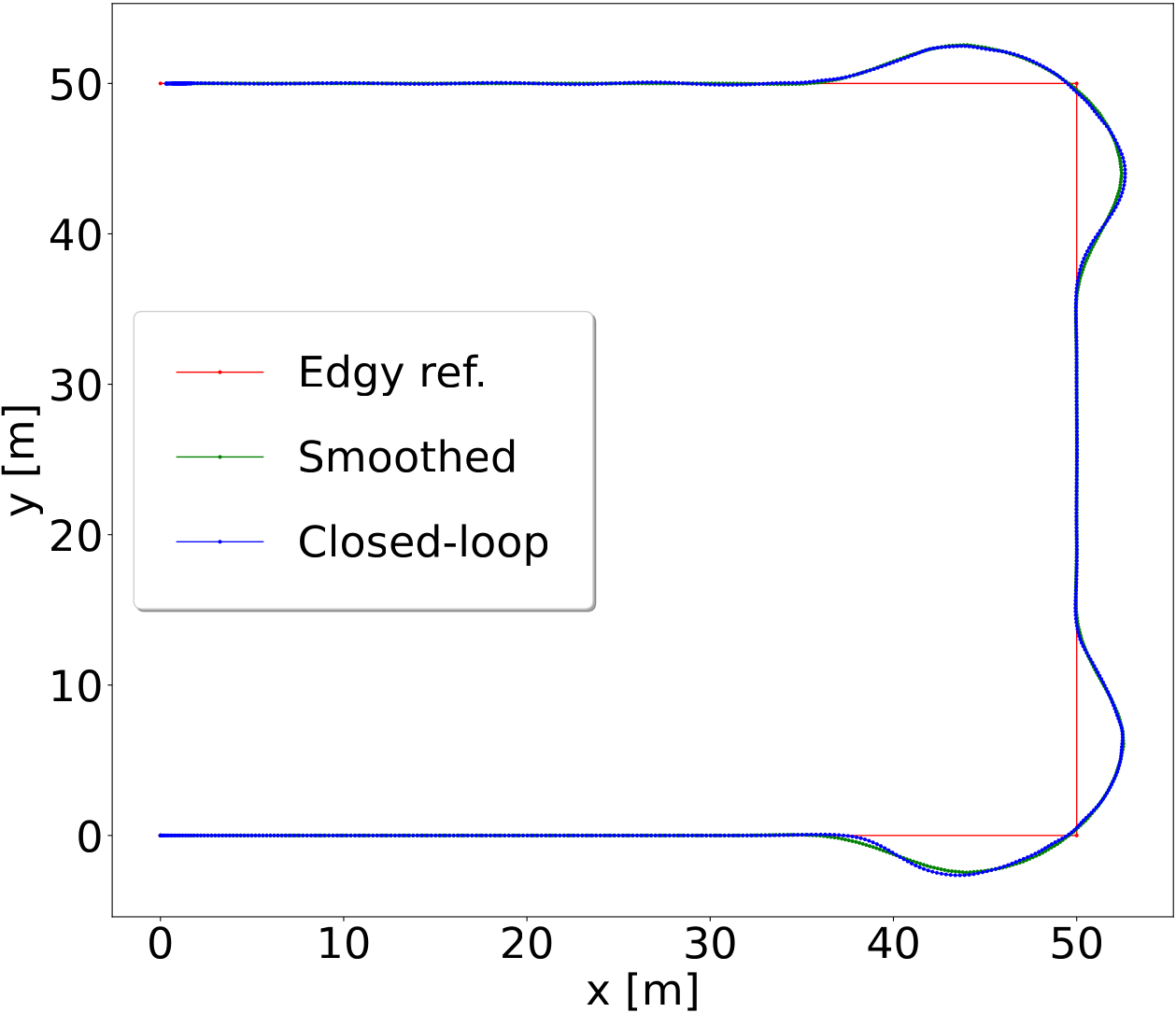}
\caption{Closed-loop tracking in $xy$-plane.\vspace*{0.1cm}}
\label{fig_ex2m1e}
\end{subfigure}
\begin{subfigure}[b]{0.239\textwidth}
\centering
\includegraphics[width=1.0\textwidth]{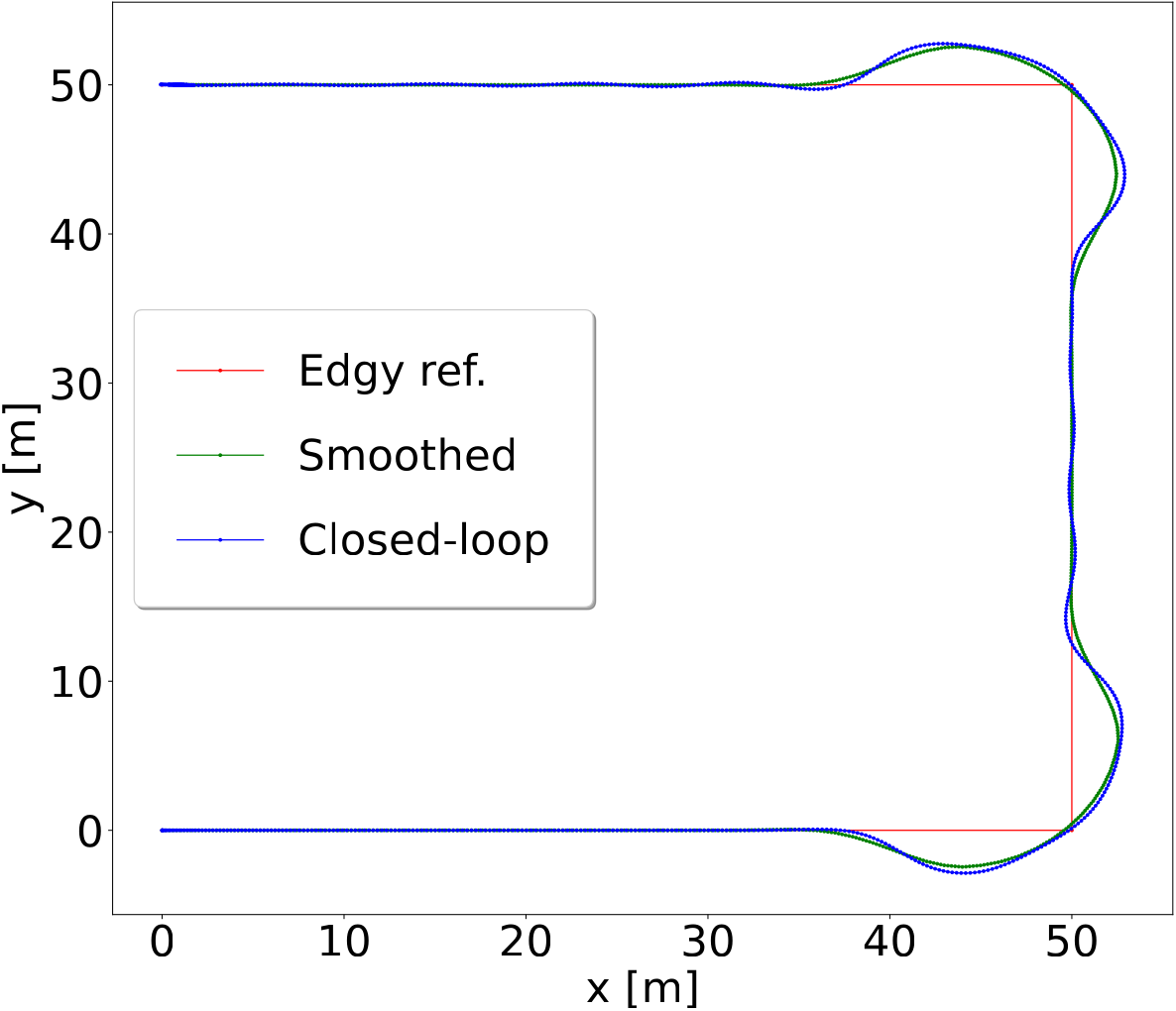}
\caption{Closed-loop tracking in $xy$-plane.\vspace*{0.1cm}}
\label{fig_ex2m1f}
\end{subfigure}
%
\caption{Results for Example 2 and $\textsf{M}_1$. The reference end position, $(x,y,z)=(0,50,10)$, is reached at approximately 45s, when the quadrotor enters a hovering state at that end position. Column-wise, the results are compared for two different versions to calculate acceleration references according to Eq. \eqref{eq_ddpd_v1} and \eqref{eq_ddpd_v2}, respectively. The latter version (see the right-column subfigures) generates much smoother control signals and less oscillations in z-direction, however at the cost of slightly more lateral overshoots with respect to the reference in the $xy$-plane.}
\label{fig_ex2_m1}
\end{figure}

\begin{figure}[htbp]
\centering
%
\begin{subfigure}[b]{0.239\textwidth}
\centering
\includegraphics[width=1.0\textwidth]{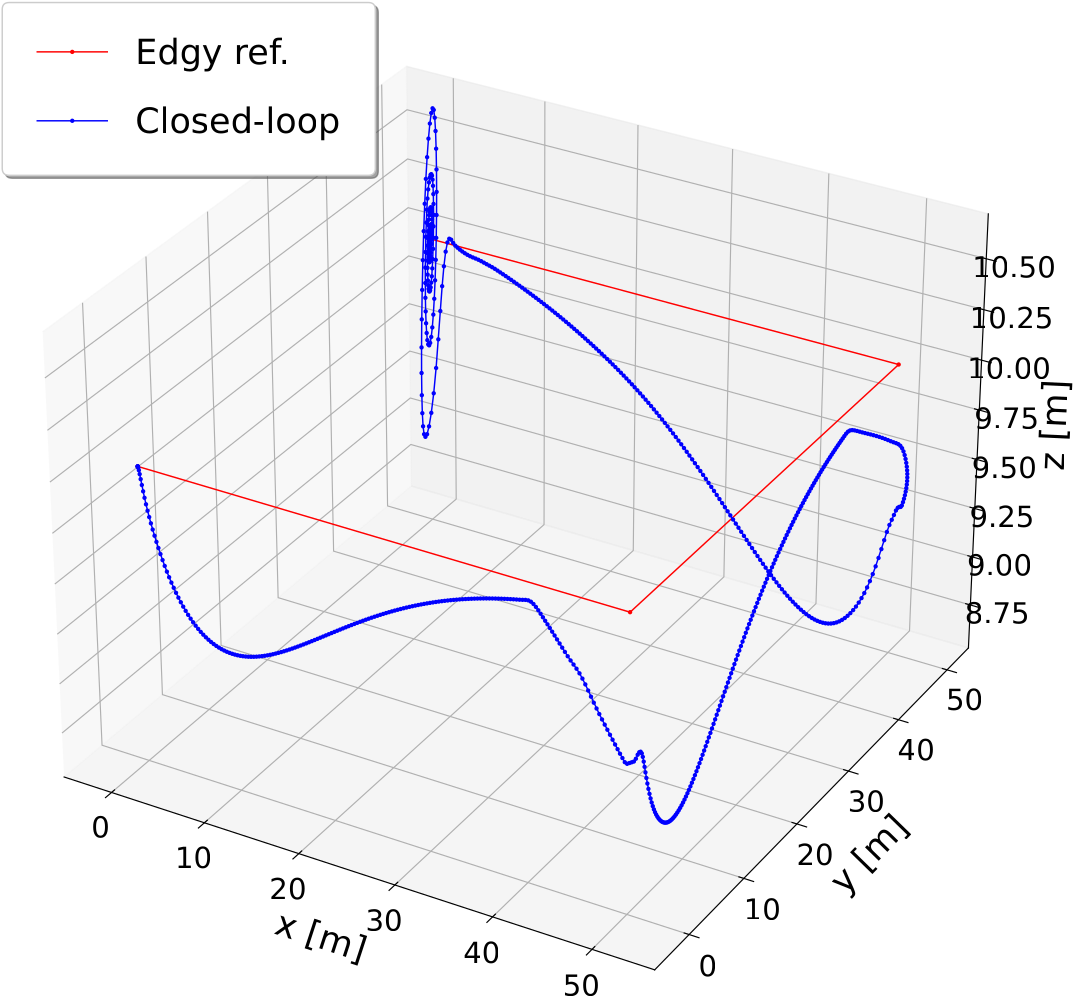}
\caption{Result in 3D. \vspace*{0.1cm}}
\label{fig_ex2m2a}
\end{subfigure}
\begin{subfigure}[b]{0.239\textwidth}
\centering
\includegraphics[width=1.0\textwidth]{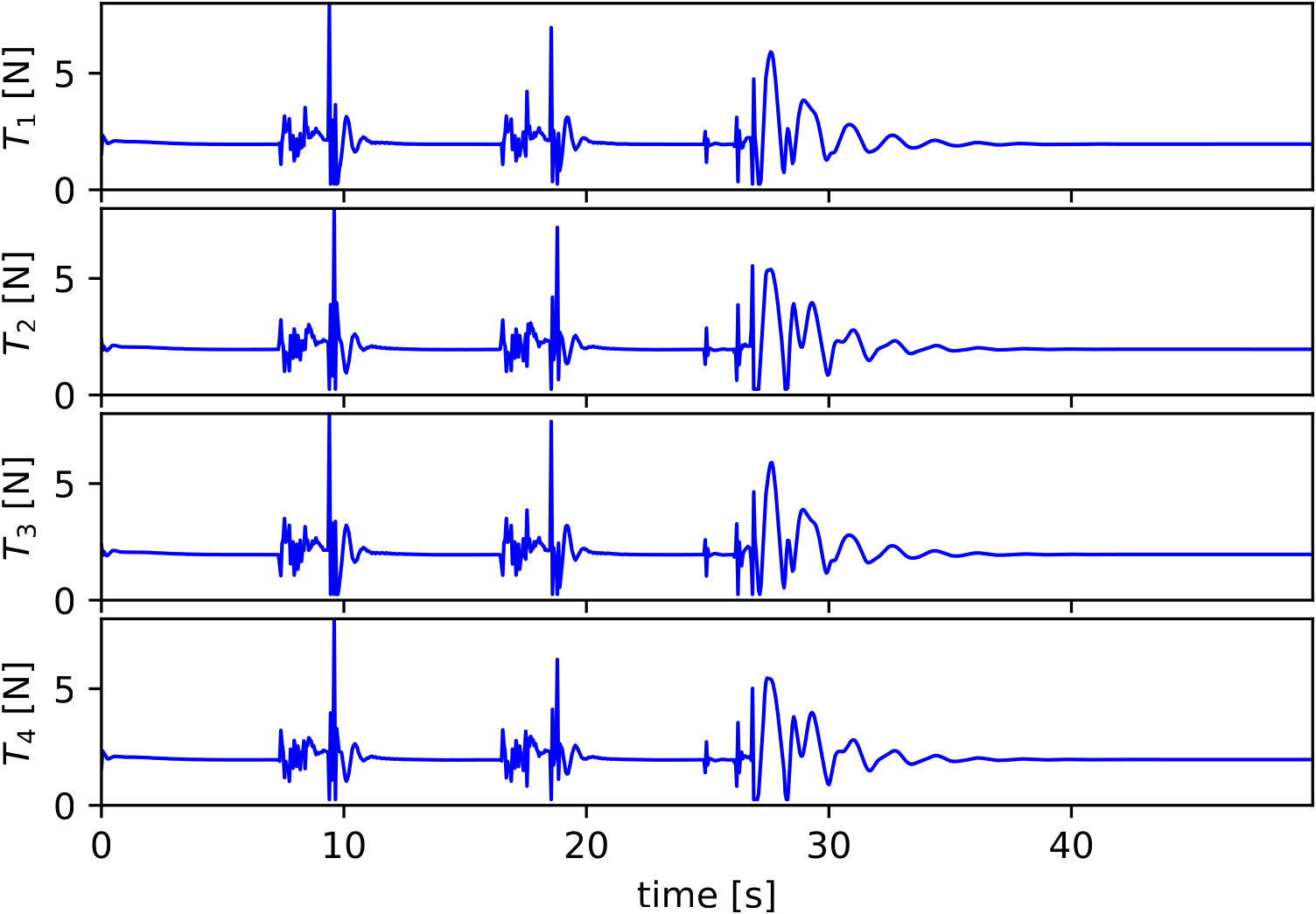}
\caption{4 control signals.\vspace*{0.1cm}}
\label{fig_ex2m2b}
\end{subfigure}
%
\begin{subfigure}[b]{0.39\textwidth}
\centering
\includegraphics[width=0.9\textwidth]{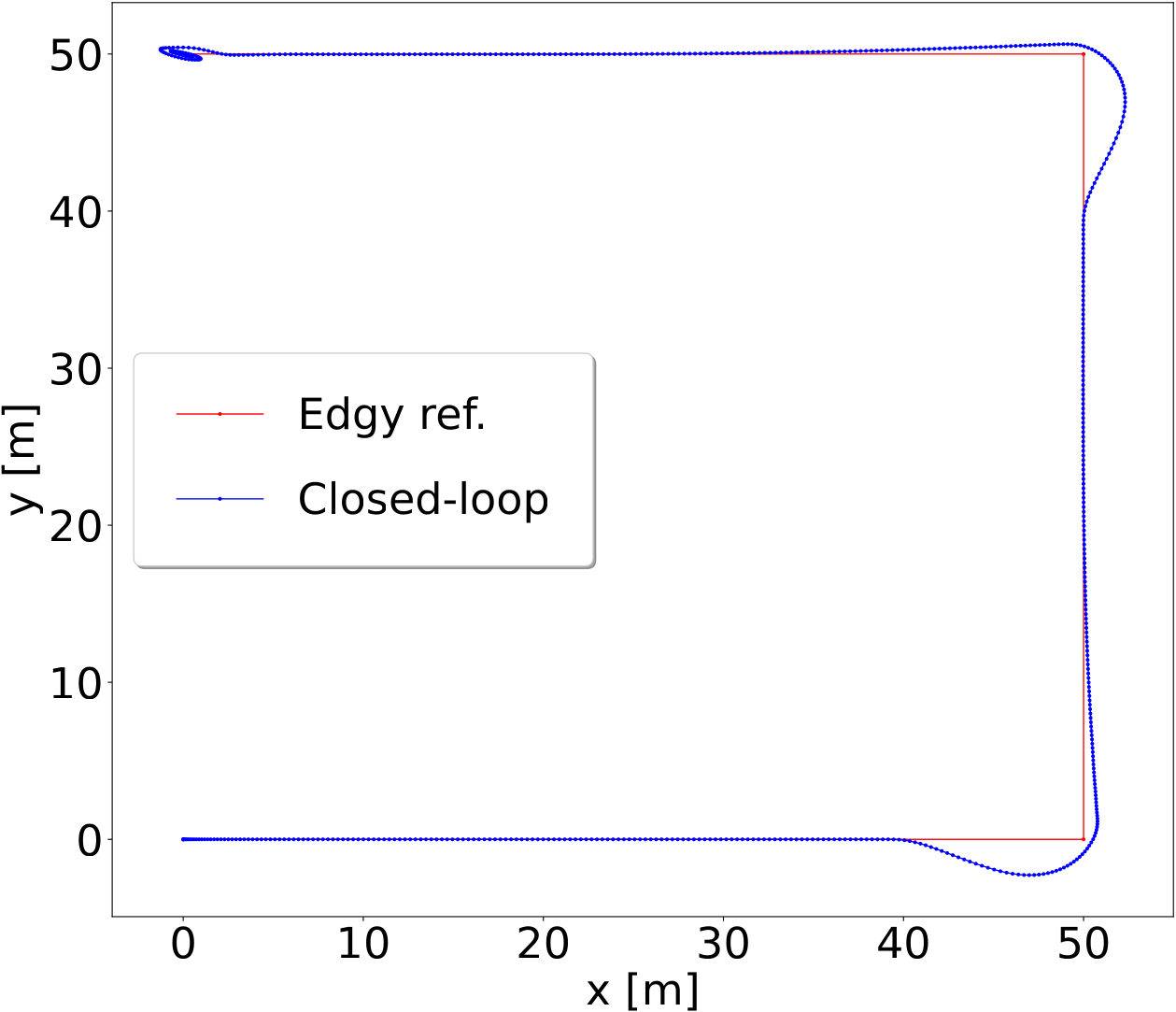}
\caption{Closed-loop tracking result in the $xy$-plane. \vspace*{0.1cm}}
\label{fig_ex2m2c}
\end{subfigure}
%
\caption{Results for Example 2 and $\textsf{M}_2$. The reference end position, $(x,y,z)=(0,50,10)$, is reached at approximately 40s, when the quadrotor enters a hovering state at that end position.}
\label{fig_ex2_m2}
\end{figure}

\begin{figure}[htbp]
\centering
%
\begin{subfigure}[b]{0.239\textwidth}
\centering
\includegraphics[width=1.0\textwidth]{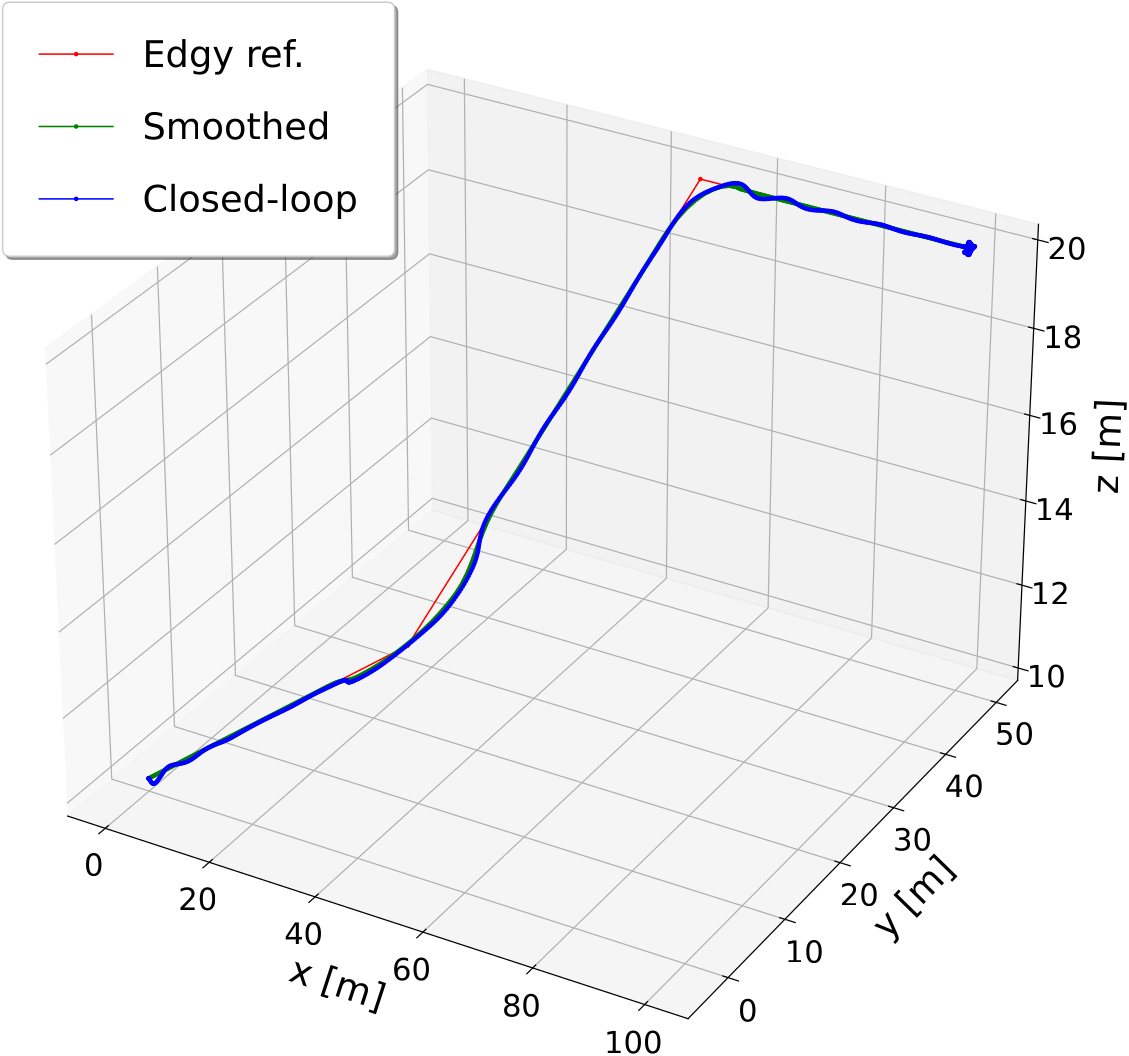}
\caption{Result in 3D. \vspace*{0.1cm}}
\label{fig_ex3m1a}
\end{subfigure}
\begin{subfigure}[b]{0.239\textwidth}
\centering
\includegraphics[width=1.0\textwidth]{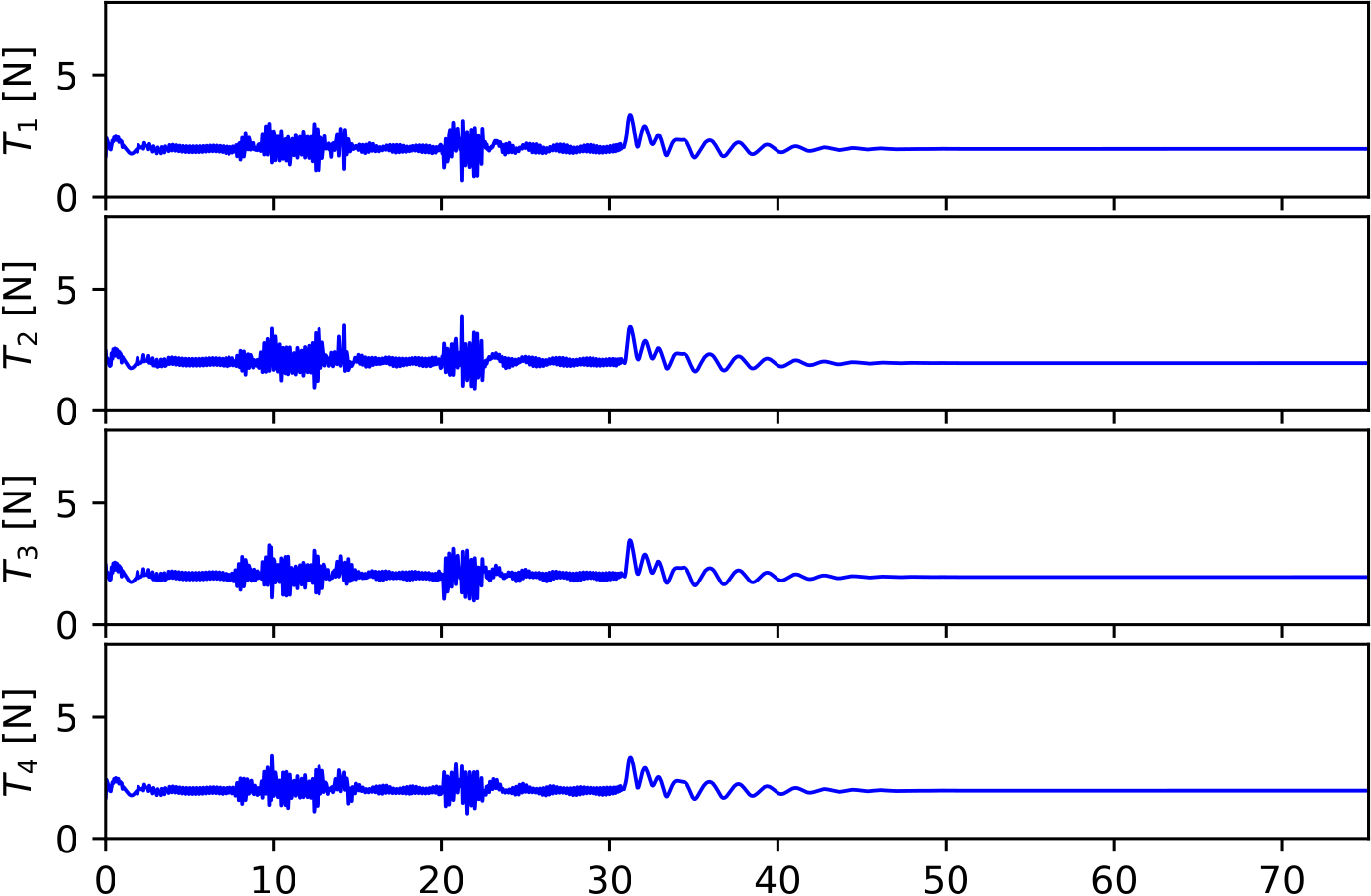}
\caption{4 control signals.\vspace*{0.1cm}}
\label{fig_ex3m1b}
\end{subfigure}
%
\begin{subfigure}[b]{0.239\textwidth}
\centering
\includegraphics[width=1.0\textwidth]{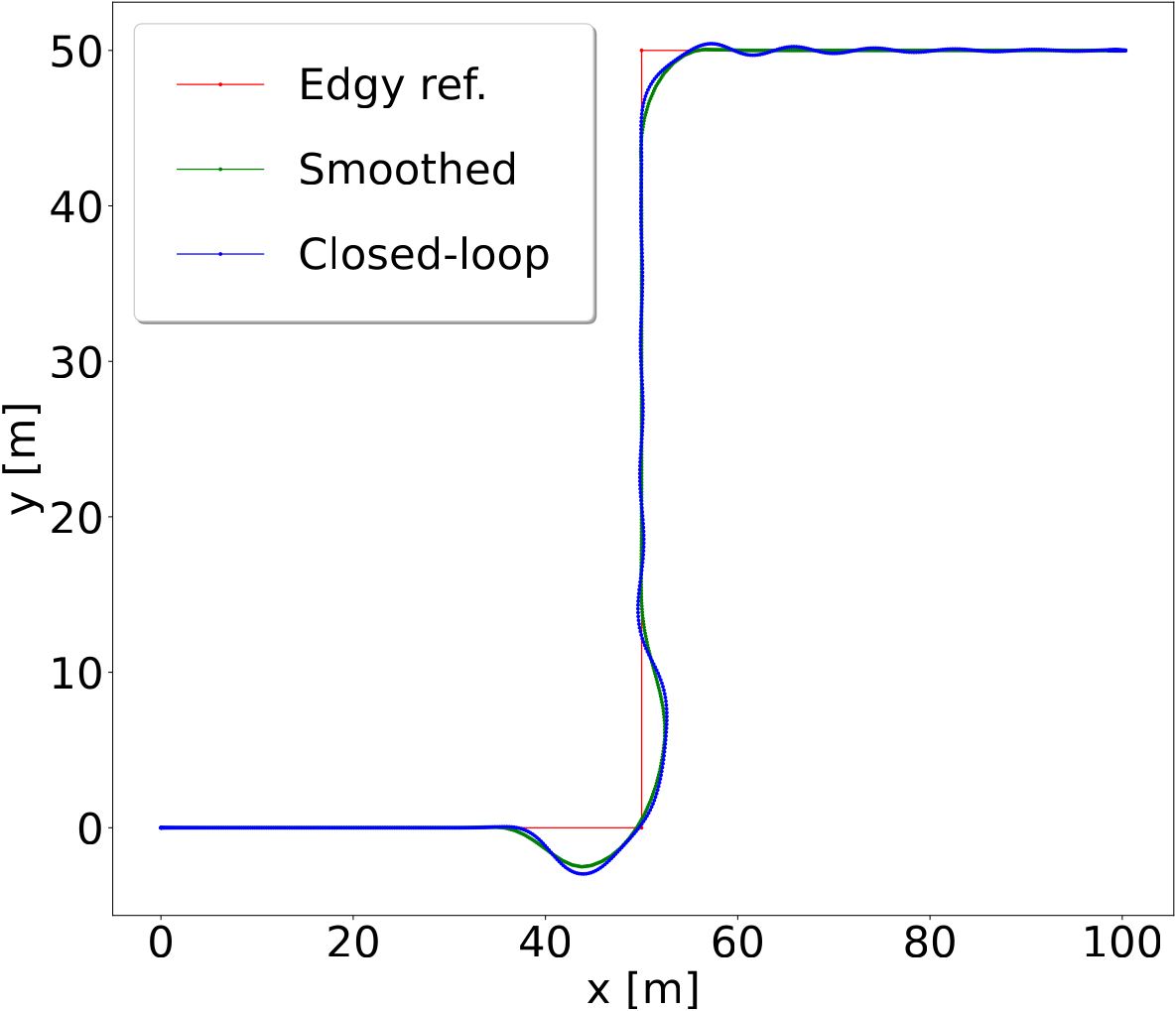}
\caption{$xy$-plane. \vspace*{0.1cm}}
\label{fig_ex3m1c}
\end{subfigure}
\begin{subfigure}[b]{0.239\textwidth}
\centering
\includegraphics[width=1.0\textwidth]{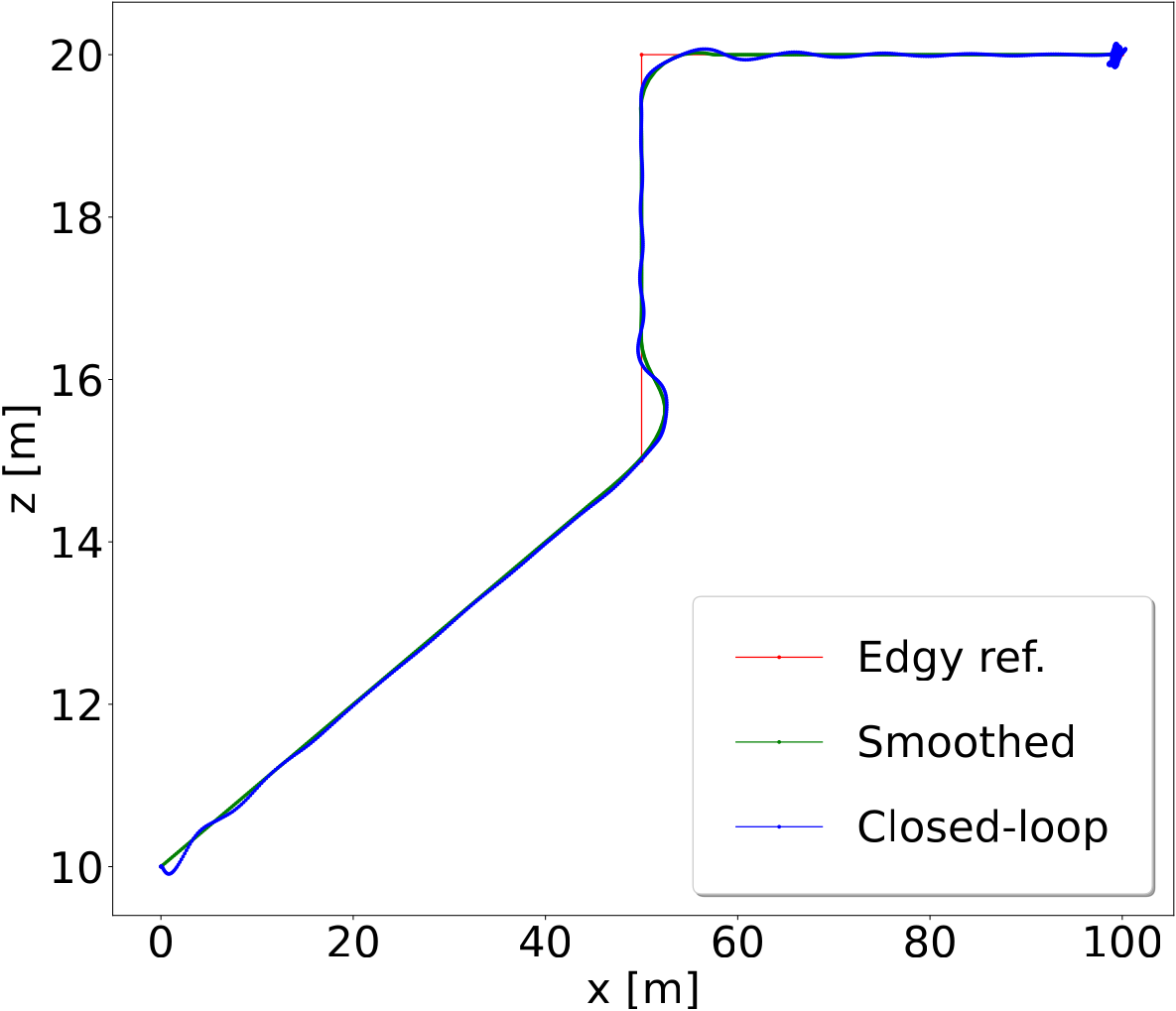}
\caption{$xz$-plane. \vspace*{0.1cm}}
\label{fig_ex3m1d}
\end{subfigure}
%
\caption{Results for Example 3 and $\textsf{M}_1$. The reference end position, $(x,y,z)=(100,50,20)$, is reached at approximately 45s, when the quadrotor enters a hovering state at that end position. The objective of staying to the right of the edgy reference within the $xy$-plane is stressed, which requires different reaching maneuvers for the left and right turns, respectively.}
\label{fig_ex3_m1}
\end{figure}

\begin{figure}
\centering
\includegraphics[width=0.8\linewidth]{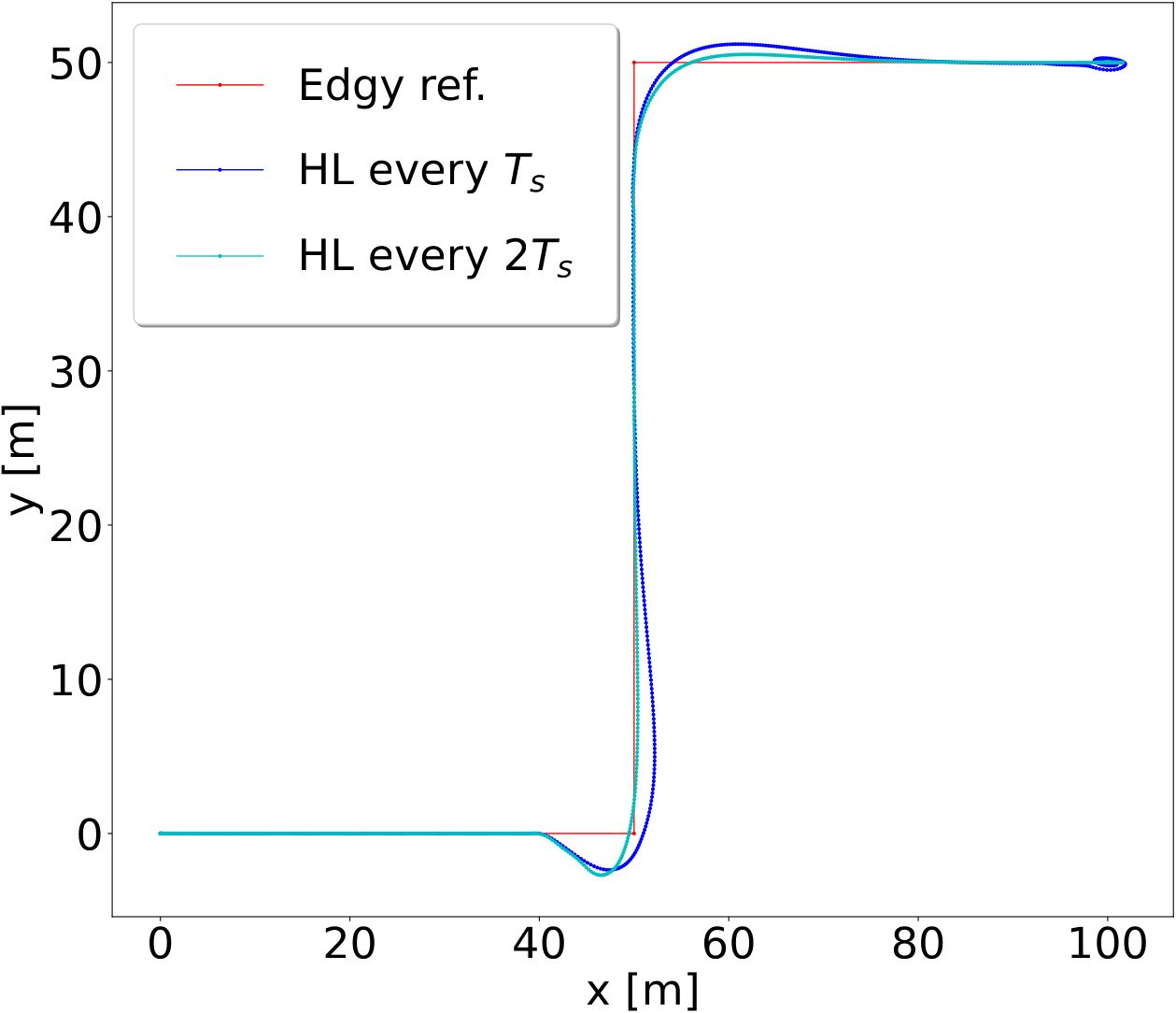}
\caption{Results for Example 3 and $\textsf{M}_2$. Comparison when high-level (HL) planning every sampling time $T_s$, or alternatively planning only every second sampling time. For the latter case, a new reference over spatial prediction horizon $H=20$m is generated every $2T_s$, and the shifted previous reference is used every other sampling time. This requires the solution of a LP only every second sampling time.}
\label{fig_every2ndTs}
\end{figure}

\begin{figure}[htbp]
\centering
%
\begin{subfigure}[b]{0.239\textwidth}
\centering
\includegraphics[width=1.0\textwidth]{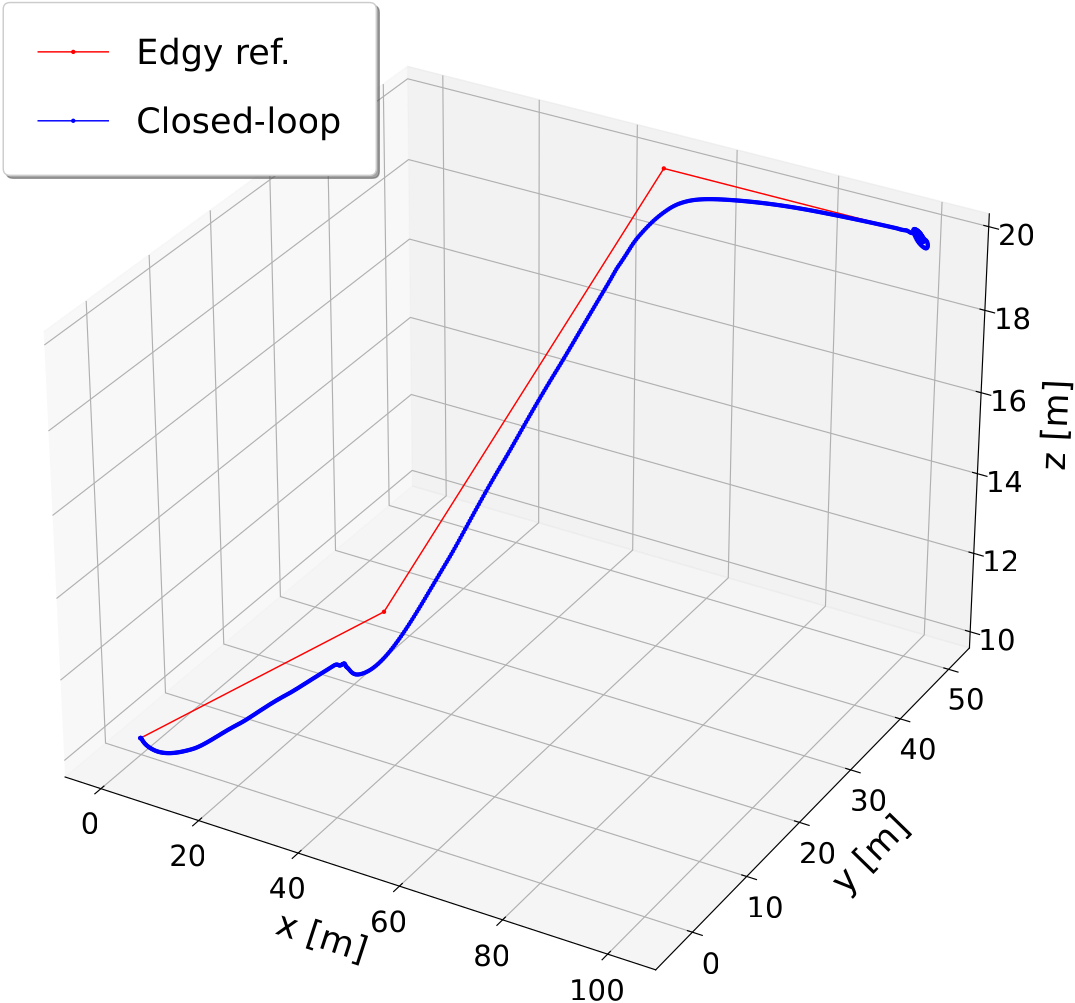}
\caption{Result in 3D. \vspace*{0.1cm}}
\end{subfigure}
\begin{subfigure}[b]{0.239\textwidth}
\centering
\includegraphics[width=1.0\textwidth]{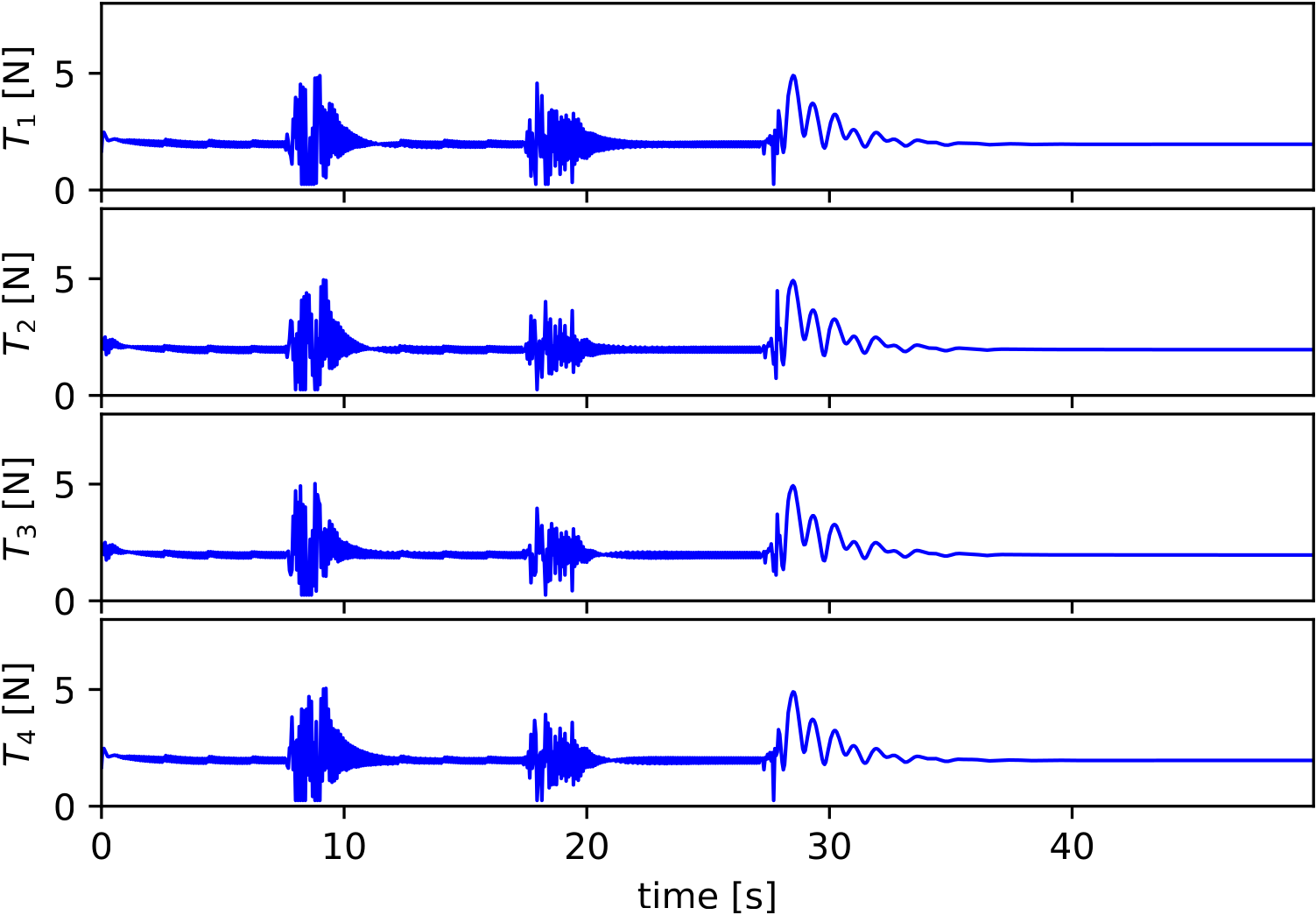}
\caption{4 control signals.\vspace*{0.1cm}}
\end{subfigure}
%
\begin{subfigure}[b]{0.239\textwidth}
\centering
\includegraphics[width=1.0\textwidth]{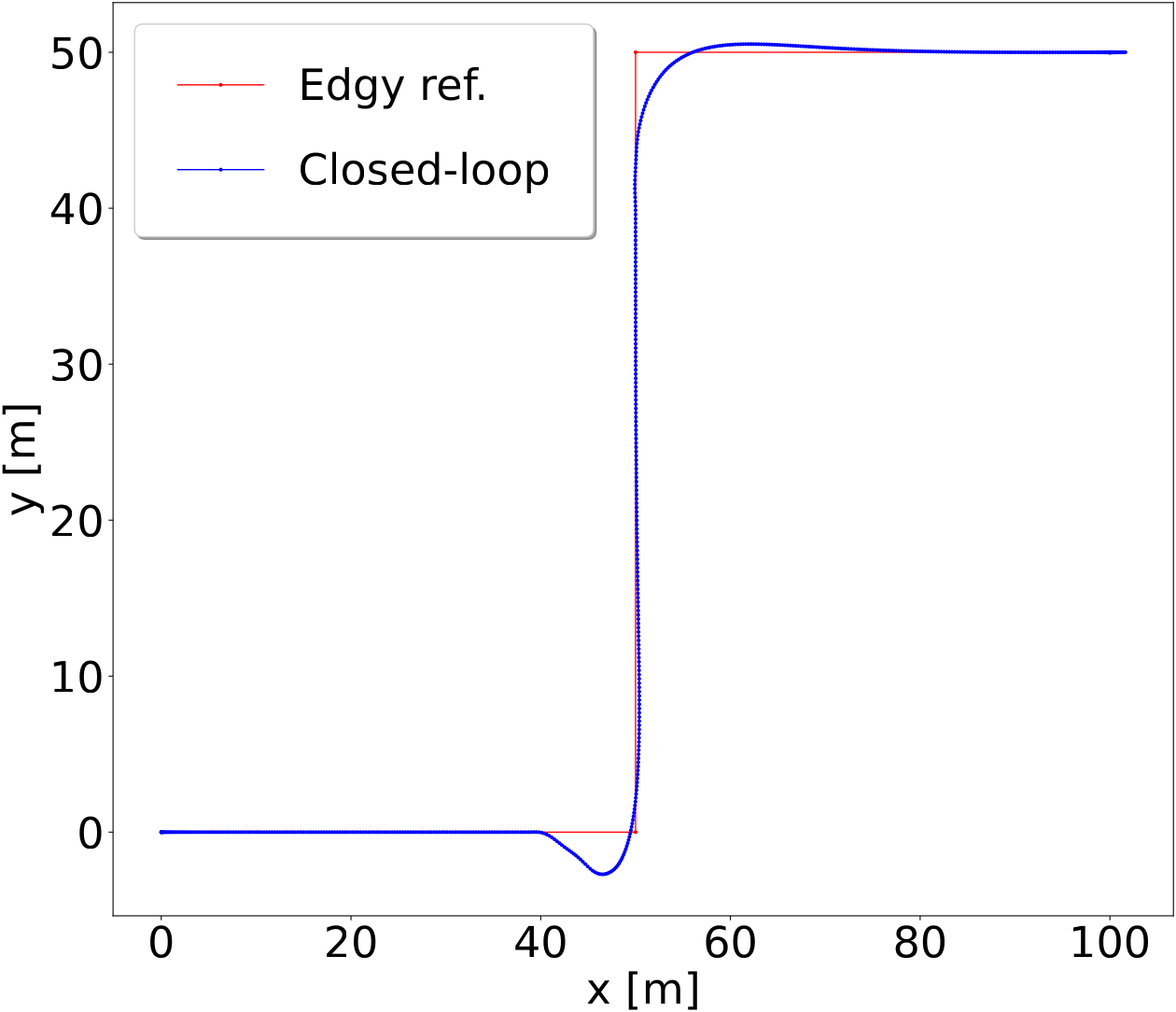}
\caption{$xy$-plane. \vspace*{0.1cm}}
\end{subfigure}
\begin{subfigure}[b]{0.239\textwidth}
\centering
\includegraphics[width=1.0\textwidth]{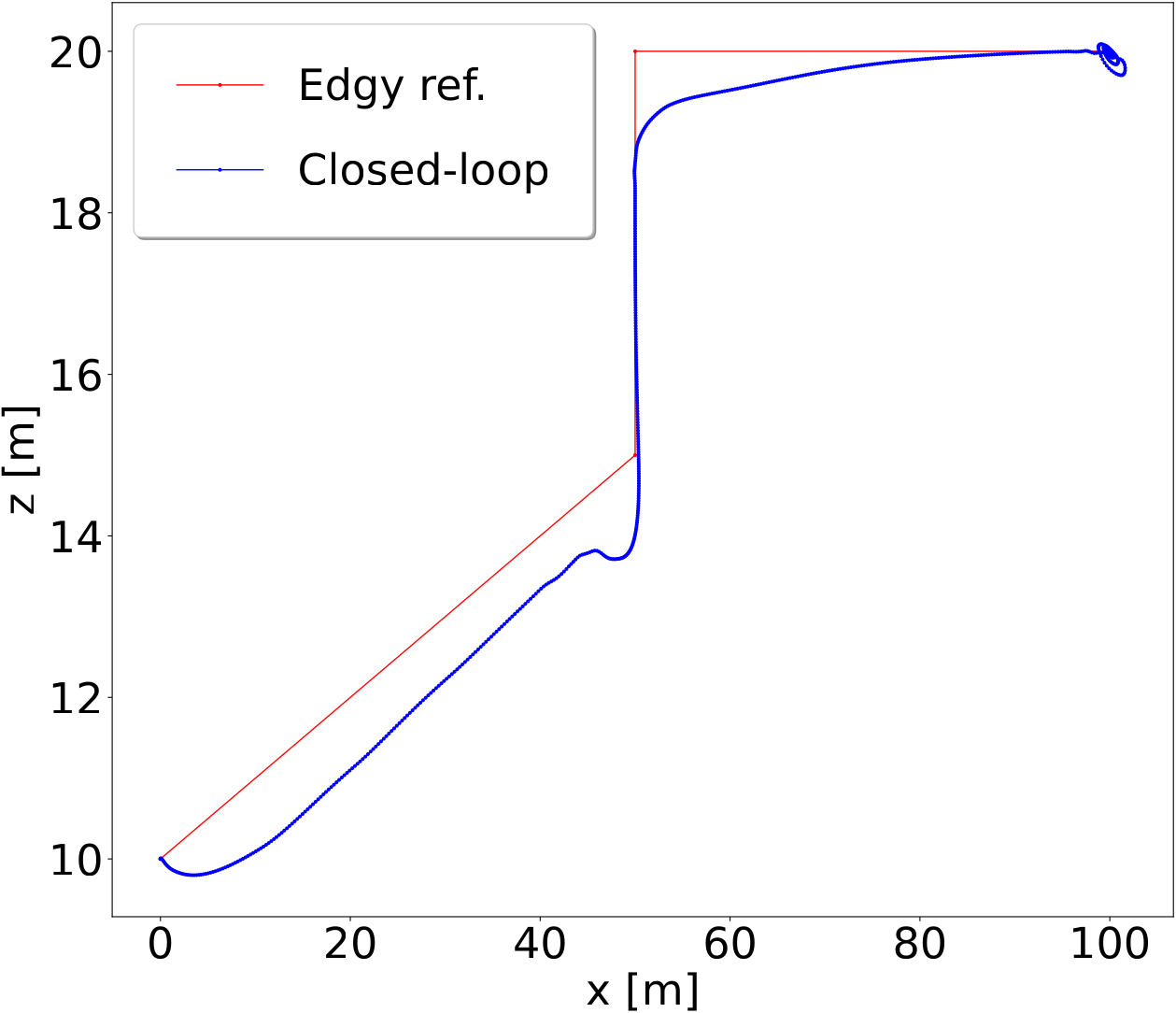}
\caption{$xz$-plane. \vspace*{0.1cm}}
\end{subfigure}
%
\caption{Results for Example 3 and $\textsf{M}_2$. A new reference over spatial prediction horizon $H=20$m is generated every $2T_s$, and the shifted previous reference is used every other sampling time. This requires the solution of a LP only every second sampling time.}
\label{fig_ex3_M2_2ndTs}
\end{figure}

Qualitative and quantitative results are summarized in Fig. \ref{fig_ex1_pid}-\ref{fig_ex3_M2_2ndTs} and Tables \ref{tab_offline}-\ref{tab_online}, respectively. One set of hyperparameters is used throughout all experiments (cf. Tab. \ref{tab_hyperparamlee}).

\begin{table}
\centering
\caption{Quantitative results for evaluation of Method $\textsf{M}_1$, where a smoothed reference is generated only once offline before it is tracked in closed-loop online. The number of state inequality constraints, number of variables, and average solve time for LP \eqref{eq_LP2} are denoted by $n_\text{p}^\text{LP}$, $n_\text{u}^\text{LP}$ and $\bar{T}_\text{avg}^\text{LP}$, respectively.\label{tab_offline}}
 \def\arraystretch{1.1}
 \begin{tabular}{|c|l|l|l|l|}
 \hline
Ex. & $n_\text{p}^\text{LP}$ & $n_\text{u}^\text{LP}$ & $\bar{T}_\text{avg}^\text{LP}$ & RMSE \\
\hline
1 & 200 & 201 & 0.007 (s) & 0.174 (m) \\
2 & 300 & 301 & 0.015 (s) & 0.160 (m) \\
3 & 302 & 303 & 0.017 (s) & 0.165 (m) \\
%
\hline 
\end{tabular}
\vspace{-0.3cm}
\end{table}


\begin{table}
\centering
\caption{Quantitative results for evaluation of Method $\textsf{M}_2$, where both a smoothed reference is generated online and tracked in closed-loop at every sampling time $T_s$. Results are for a spatial prediction horizon of 20m, interpolation spacing 0.2m, $n_\text{p}^\text{LP}=202$ and $n_\text{u}^\text{LP}=203$. \label{tab_online}}
 \def\arraystretch{1.1}
 \begin{tabular}{|c|l|}
 \hline
Ex. & $\bar{T}_\text{avg}^\text{LP}$ \\
\hline
1 & 0.0079 (s) \\
2 & 0.0078 (s) \\
3 & 0.0076 (s) \\
%
\hline 
\end{tabular}
\vspace{-0.3cm}
\end{table}

Several comments can be made. For Example 1 and $\textsf{M}_1$ two comparisons are made. First, using an EMA-filter resulted in RMSE=0.192m. In contrast, using the SG-filter resulted in RMSE=0.174m, which is 9\% lower. Throughout experiments it was observed that the SG-filter performed slightly better, however the EMA-filter offers the benefit of being simpler (algebraic).  


Second, using an alternative cascaded PID-controller \cite{lopez2023pid,mellinger2012trajectory} for low-level tracking with a cascade of positition, attitude and body angulare rate control loops resulted in RMSE=0.324m and RMSE=0.356m for the SG- and EMA-filters, respectively. These RMSEs are approximately twice as high in comparison to using the geometric controller. Qualitative results are visualized in Fig. \ref{fig_ex1_pid}. Throughout experiments for low-level tracking better performance was observed for the geometric controller, which also offers theoretical guarantees \cite{lee2010geometric}. However, the performance of the geometric controller strongly relies on feedforward terms which are a function of suitably smoothed references. A cascade of 3 PID controllers was used, whereby the integral term was set zero in all 3 cases and the derivative term is set zero in 2 cases, such that a total of 4 diagonal $3 \times 3$ tuning matrices result. This procedure was done for comparison reasons such that the number of tuning matrices was equal to the 4 diagonal tuning matrices required for the geometric controller. 

For Example 1 and $\textsf{M}_2$ two spatial prediction horizons were compared. Using a reduced spatial prediction horizon of 10m, instead of 20m, resulted in the number of LP-constraints $n_\text{cstrts}^\text{LP}=102$, the number of optimization variables $n_\text{var}^\text{LP}=103$ and solve time $\bar{T}_\text{avg}^\text{LP}=0.003$s, which is 62.5\% lower than $\bar{T}_\text{avg}^\text{LP}=0.008$s in Table \ref{tab_online}. However, due to the reduced horizon the closed-loop tracking performance is reduced and results in delayed (less anticipative) reaching maneuvers. See Fig. \ref{fig_influenceH} for illustration.

While for $\textsf{M}_1$ large spatial interpolation spacing $D_s=1$m could be used for the solution of LP \eqref{eq_LP2}, for $\textsf{M}_2$ much smaller distances had to be selected with $D_s=0.2$m. The reason is the receding reference generation method illustrated in Fig. \ref{fig_refgen_M2}, which required smaller spacing for sufficient accuracy. 

Fig. \ref{fig_ex2_m1} demonstrates the sensitivity of the geometric controller to different feedforward terms calculation methods.

For Example 3 and $\textsf{M}_2$, an overshooting behavior could be observed, which does not occur for $\textsf{M}_1$ for the reference smoothing step. The reason for this behavior is the receding reference generation step and the limited spatial prediction horizon $H$. One method to mitigate this effect is to not replan a reference trajectory (over prediction horizon $H$) at every sampling time, but instead replan only every second sampling time, and use the shifted previous plan every other sampling time. The problem complexity remains the same, however, a linear program needs now to be solved only every second sampling time. See Fig. \ref{fig_every2ndTs} for the effect.

Overall, all experiments could be solved as illustrated for both the $\textsf{M}_1$ and $\textsf{M}_2$-setting, validating the general approach of combining high-level spatial Dubins airplane-based reference smoothing with low-level geometric tracking.

\section{Discussion, limitations and outlook\label{sec_discussion}}

The present paper proposes high-level reference path smoothing based on a lower-dimensional Dubins airplane model. In combination with the decoupling and spatial modeling approach from \cite{plessen2026simple} this yields a small LP with (i) only 1 variable per discretization step in the formulation of \eqref{eq_LP2}, and with (ii) the additional advantage that its objective function \eqref{eq_LP2_objFcn} is hyperparameter-free.

Alternatively, instead of the Dubins airplane model the full quadrotor model might be used for the formulation of an optimization problem for the tracking of references subject to lateral constraints. This is not trivial. It would increase the state-space to 12 states (or 13 in case of quaternions), and the input-space to 4 controls per discretization step. An initialization (or the initial references in case of successive linearization for nonlinear model handling) of an optimal control problem for such a large state-input space is not obvious, in particular for the tracking of edgy reference paths. Furthermore, a Frenet-Serret frame may be introduced which would introduce singularities requiring careful handling \cite{krinner2024mpcc++,spedicato2017minimum}. Such a comparison is left for future work.

\section{Conclusion\label{sec_conclusion}}

A method for quadrotor control was presented. It combines a high-level reference smoothing step with a low-level reference tracking step. The high-level step is based on a low-dimensional Dubins airplane model and leverages its structure for decoupling, spatial modeling and the formulation of a small linear program. The low-level step leverages a geometric tracking controller, which is based on the full quadrotor model. It was differentiated between two different setups. Either the high-level planning step is conducted once and offline, or, alternatively, the high-level planning step is conducted recedingly online and over a limited spatial prediction horizon.

It was found that proposed two-level approach permits to smooth edgy reference paths over large planning horizons at low computational cost, while offering reference tracking as well as lateral trajectory shaping ability. Heuristics for reference velocity shaping and a filtering step for additional smoothness significantly improved performance.





Future work will attempt to apply the same methodology to land- and sea-based path planning problems, where it is expected that only the low-level (algebraic) controller has to be exchanged depending on the application.

\nocite{*}
\bibliographystyle{ieeetr}
\bibliography{myref}

\end{document}